\documentclass[default,iicol]{sn-jnl}
 
\usepackage{natbib}
\setcitestyle{authoryear,open={(},close={)}}
\usepackage{bm}
\usepackage{subfig}
\usepackage{commath}
\usepackage{caption}
 
\jyear{2022}%
 
\theoremstyle{thmstyleone}%
\theoremstyle{thmstyletwo}%

\theoremstyle{thmstylethree}%

\begin{document}
 
\title[Article Title]{ResNet Structure Simplification with the Convolutional Kernel Redundancy Measure}
 
 
\author*[1,2]{\fnm{Hongzhi} \sur{Zhu}}\email{hzhu@ece.ubc.ca}
 
\author[1,2,3]{\fnm{Robert} \sur{Rohling}}\email{rohling@ece.ubc.ca}
 
\author[1,2]{\fnm{Septimiu} \sur{Salcudean }}\email{tims@ece.ubc.ca}
 
\affil*[1]{\orgdiv{Department of Electrical and Computer Engineering}, \orgname{University of British Columbia}, \orgaddress{\street{5500-2332 Main Mall}, \city{Vancouver}, \postcode{V6T 1Z4}, \state{BC}, \country{Canada}}}
 
\affil[2]{\orgdiv{School of Biomedical Engineering}, \orgname{University of British Columbia}, \orgaddress{\street{251-2222 Health Sciences Mall}, \city{Vancouver}, \postcode{V6T 1Z4}, \state{BC}, \country{Canada}}}
 
\affil[3]{\orgdiv{Department of Mechanical Engineering}, \orgname{University of British Columbia}, \orgaddress{\street{2054-6250 Applied Science Lane}, \city{Vancouver}, \postcode{V6T 1Z4}, \state{BC}, \country{Canada}}}
 
 
\abstract{Deep learning, especially convolutional neural networks, has triggered accelerated advancements in computer vision, bringing changes into our daily practice. Furthermore, the standardized deep learning modules (also known as backbone networks), i.e., ResNet and EfficientNet, have enabled efficient and rapid development of new computer vision solutions. Yet, deep learning methods still suffer from several drawbacks. One of the most concerning problems is the high memory and computational cost, such that dedicated computing units, typically GPUs, have to be used for training and development. Therefore, in this paper, we propose a quantifiable evaluation method, the convolutional kernel redundancy measure, which is based on perceived image differences, for guiding the network structure simplification. When applying our method to the chest X-ray image classification problem with ResNet, our method can maintain the performance of the network and reduce the number of parameters from over $23$ million to approximately $128$ thousand (reducing $99.46\%$ of the parameters).}

 
 
 
\maketitle
 
\section{Introduction}\label{sec:Introduction}
 
The human visual system is known for its wide adaptivity under very different illumination conditions, and sharp acuity despite the minute feature contrasts, especially of the trained professionals \citep{zheng2020assessment}. This intelligent system has inspired the development of modern computer vision (CV) solutions, including deep learning (DL), that can sometimes outperform expert level performance.
\par
Traditionally, engineered features are used for the detection and extraction of related information from images for a designated task. Some of the frequently used features include scale-invariant feature transform (SIFT) \citep{lowe1999object}, and histogram of oriented gradients (HOG) \citep{dalal2005histograms}, and they have demonstrated a high degree of robustness in various CV tasks, i.e., detection, and image recognition. Still, convolutional neural network (CNN) methods, also known as space invariant artificial neural networks, demonstrate superior performance thanks to the following three characteristics. The first characteristic lies in its representative learning capability, such that features used for information extraction are no longer designed with engineered efforts, but based on the training dataset, formulated automatically during the training processes \citep{zhang2018network, bengio2013representation}. This enables the CNN method to be more adaptive to a much broader range of CV tasks, and saves human labour for dedicated feature design. The second characteristic is the ensemble of features at different levels of abstraction. While traditional methods, e.g., SIFT and HOG, mostly rely on one set of features for information extraction, CNN methods typically employ hierarchical sets of features that enable more complicated and abstracted information compounding \citep{yamashita2018convolutional}. Therefore, a deeper CNN generally has greater information abstraction capability than a shallow CNN, with the side effect that the deeper features are increasingly harder to interpret. The lack of interpretability is one of the most criticized characteristics of the CNN method, despite some efforts to explain how a particular CNN works \citep{zhang2018interpretable, zhu2022gaze, selvaraju2017grad}. The third characteristic is over-parameterization, where, depending on the task, a huge portion of trainable parameters in the network may be redundant \citep{soltanolkotabi2018theoretical}. While some argue that over-parameterization is the essence of the CNN method \citep{allen2019convergence, arora2018optimization}, it also brings the problems of high memory and computational cost.
\par
In our earlier study, we proposed the gaze-guided class activation mapping (GG-CAM) method \citep{zhu2022gaze}, which advances the CNN interpretability problem with the help of gaze tracking technology. In this paper, we focus on the problem of high memory and computational cost, and propose a novel measurement to estimate the redundancy of convolution kernels in the network from which we can optimize a network with reduced trainable parameters.
\par
There mainly exist three streams of solutions to reduce the memory and computational cost for CNN methods. The first stream is called knowledge distillation, where we start by training a large (memory costly and computationally heavy) network, and then try to transfer the knowledge learnt from the large network to a lighter network that is much less memory costly and computationally heavy \citep{gou2021knowledge}. However, knowledge distillation is procedurally intricate, and cannot guarantee a desired outcome \citep{cho2019efficacy}. The second stream consists of development of light CNN structures with computation optimization, i.e., MobileNet \citep{howard2017mobilenets} and ShuffleNet \citep{zhang2018shufflenet}. Still, light CNN structures may fail to achieve a comparable performance as the conventional CNN structures, i.e., ResNet and EfficientNet. The last stream of method is through network structure optimization. When the network structures' are parameterized, one can adjust the structure of the networks for desired performance boosts and/or computational load reduction via automated structure search \citep{xue2021multi} or through manual tuning \citep{d2020structural, wu2019resnet}. The new method proposed in this paper also falls into this category, with the major advantage that it can directly guide the structure parameter tuning process, both visually and quantitatively, such that we can complete the tuning process with higher efficiency and confidence.
\par
To demonstrate our method, the rest of the paper is organized as follows. We start by formulating our objectives in Section \ref{sec:Problem Formulation}, where more related background is also discussed. Following that, we propose our method in Section \ref{sec:Solution}. To evaluate our method, experiment results with visualization are presented in Section \ref{sec:Evaluation and Visualization}. We conclude this paper in Section \ref{sec:Conclusion}.
 
 
\begin{table*}[th]
\centering
\begin{tabular}{@{}c|ccccc@{}}
\toprule
 
 \includegraphics[width=0.1\textwidth]{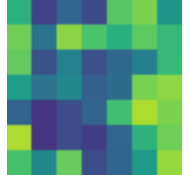}&
\includegraphics[width=0.1\textwidth]{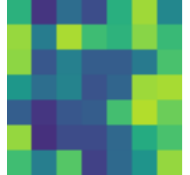}& \includegraphics[width=0.1\textwidth]{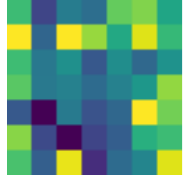}& \includegraphics[width=0.1\textwidth]{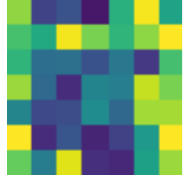}& \includegraphics[width=0.1\textwidth]{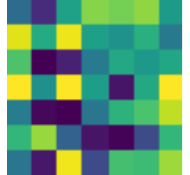}& \includegraphics[width=0.1\textwidth]{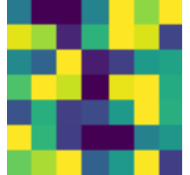} \\ 
 $\bm{x}$&
 $\bm{y}_{1,1}$=$\bm{x}+\sigma_{1}$&
 $\bm{y}_{1,2}$=$\bm{x}+\sigma_{2}$&
 $\bm{y}_{1,3}$=$\bm{x}+\sigma_{3}$&
 $\bm{y}_{1,4}$=$\bm{x}+\sigma_{4}$&
 $\bm{y}_{1,5}$=$\bm{x}+\sigma_{5}$\\ \midrule
 $\psi_{1,1,1}$&
 0.987& 0.867& 0.822& 0.532& 0.375\\
 $\psi_{0.1,1,1}$&
 0.987& 0.867& 0.824& 0.532& 0.379
 \\ \midrule
 &
\includegraphics[width=0.1\textwidth]{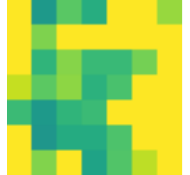}& \includegraphics[width=0.1\textwidth]{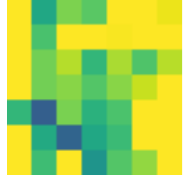}& \includegraphics[width=0.1\textwidth]{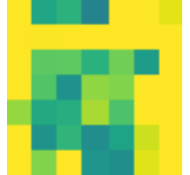}& \includegraphics[width=0.1\textwidth]{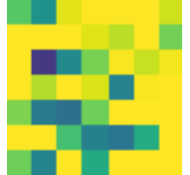}& \includegraphics[width=0.1\textwidth]{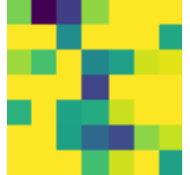}\\
 &
 $\bm{y}_{2,1}$=$\bm{y}_{1,1}+0.5$&
 $\bm{y}_{2,2}$=$\bm{y}_{1,2}+0.5$&
 $\bm{y}_{2,3}$=$\bm{y}_{1,3}+0.5$&
 $\bm{y}_{2,4}$=$\bm{y}_{1,4}+0.5$&
 $\bm{y}_{2,5}$=$\bm{y}_{1,5}+0.5$\\ \midrule
 $\psi_{1,1,1}$&
 0.803& 0.700& 0.656& 0.428& 0.294\\
 $\psi_{0.1,1,1}$&
 0.967& 0.849& 0.806& 0.521& 0.369\\
 \bottomrule
 
\end{tabular}
\caption{Visual demonstration of difference similarity measures. The reference image patch $\bm{x}$ has a dimension of $7\times7$. $\sigma_i\in\mathbb{R}^{7\times7}$ represents additive random Gaussian noise, which each element in $\sigma_i$ is independently drawn from Gaussian distribution $\mathcal{N}(0, (i/10-0.05)^2))$, $i=1,2,...,5$. From left to right, the noise level increases. From the top row image to the corresponding bottom row image, a constant value of $0.5$ is added. Both $\psi_{1,1,1}$ and $\psi_{0.1,1,1}$ quantities are presented to compare how similarity measurement changes when we perform different transforms to $x$.}
\label{tab:demo}
\end{table*}
 
\begin{table*}[th]
\centering
\begin{tabular}{@{}c|c|c@{}}
\toprule
 Convolution Layers & Optimized ResNet50 & Standard ResNet50 \\ \midrule
 ${1}$ & $16\times 1\times 7\times 7$ & $64\times 1\times 7\times 7$\\
 ${2,3}$ & $16\times 16\times 3\times 3$ & $64\times 64\times 3\times 3$\\
 ${4,5,6,7}$ & $32\times 32\times 3\times 3$ & $128\times 128\times 3\times 3$\\
 ${8, 9, 10, 11, 12, 13}$ & $24\times 24\times 3\times 3$ & $256\times 256\times 3\times 3$\\
 ${14, 15, 16, 17}$ & $18\times 18\times 3\times 3$ & $512\times 512\times 3\times 3$\\ \midrule
 Total number of parameters & 128062 & 23534467\\
 \bottomrule
\end{tabular}
\caption{Comparison between optimized ResNet50 and standard ResNet50 structures. The dimensions of the $17$ convolution kernels in ResNet are presented. $1\times 1$ convolution layers (linear layers), batch normalization layers, pooling layers, concatenation layers, and activation layers are not presented in this table, and their dimension can be easily inferred from the convolution kernels shown in this table.}
\label{tab:structure}
\end{table*}
 
\begin{table*}[ht]
\centering
\begin{tabular}{@{}cc|cc@{}}
\toprule
 \multicolumn{2}{c|}{Optimized ResNet50} & \multicolumn{2}{c}{Standard ResNet50}\\ \midrule
 \includegraphics[width=0.200\textwidth]{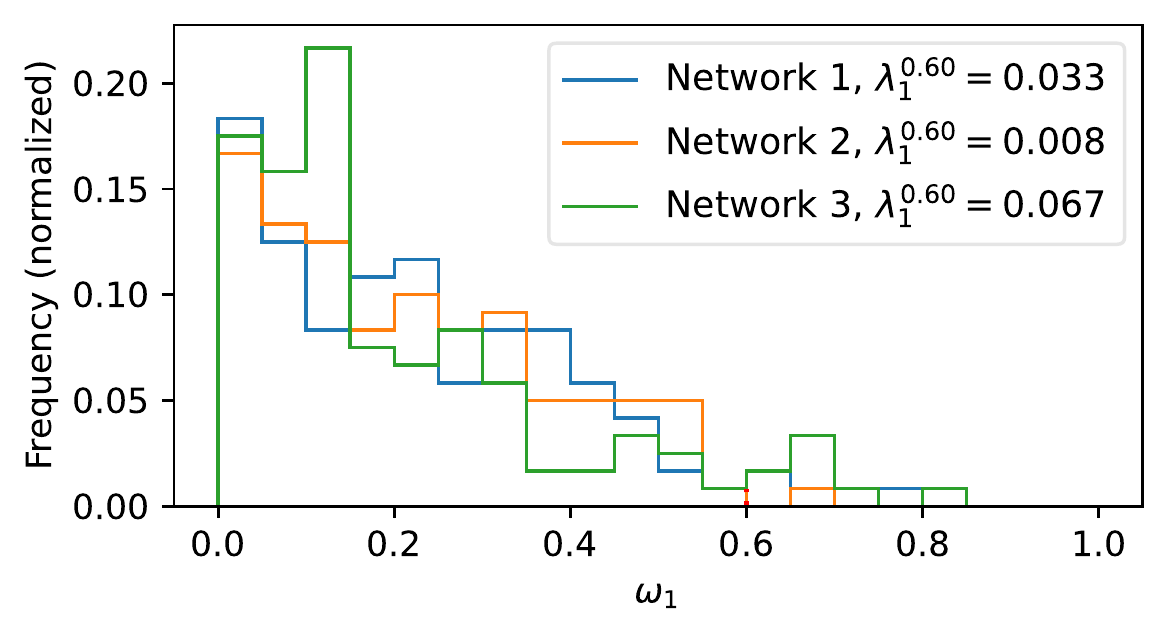}& \includegraphics[width=0.200\textwidth]{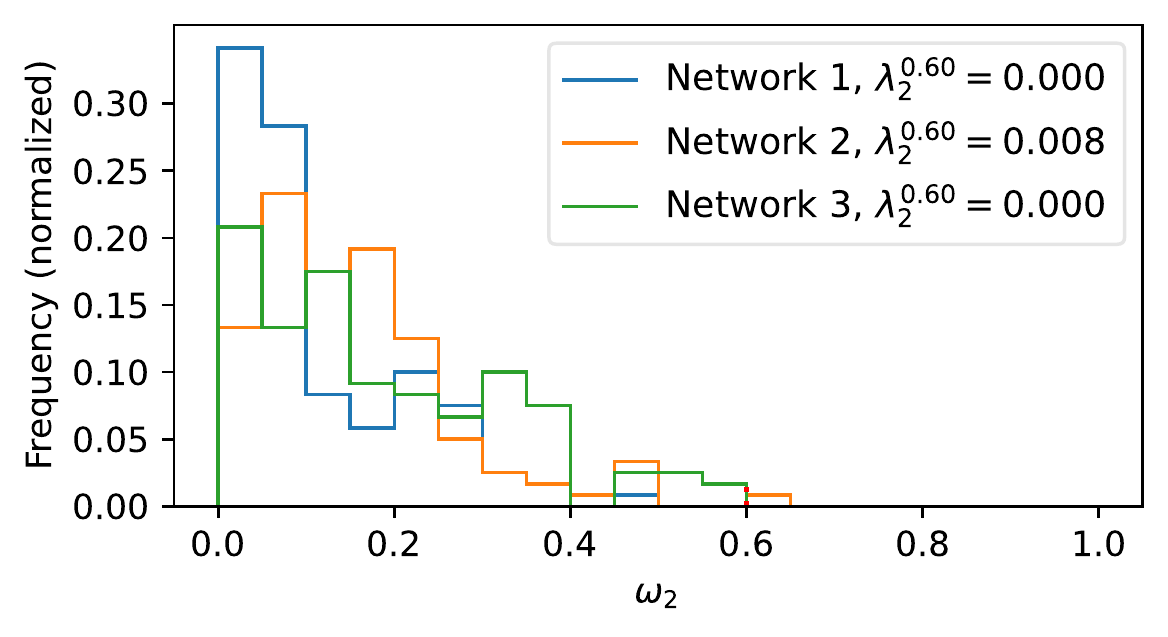} & \includegraphics[width=0.200\textwidth]{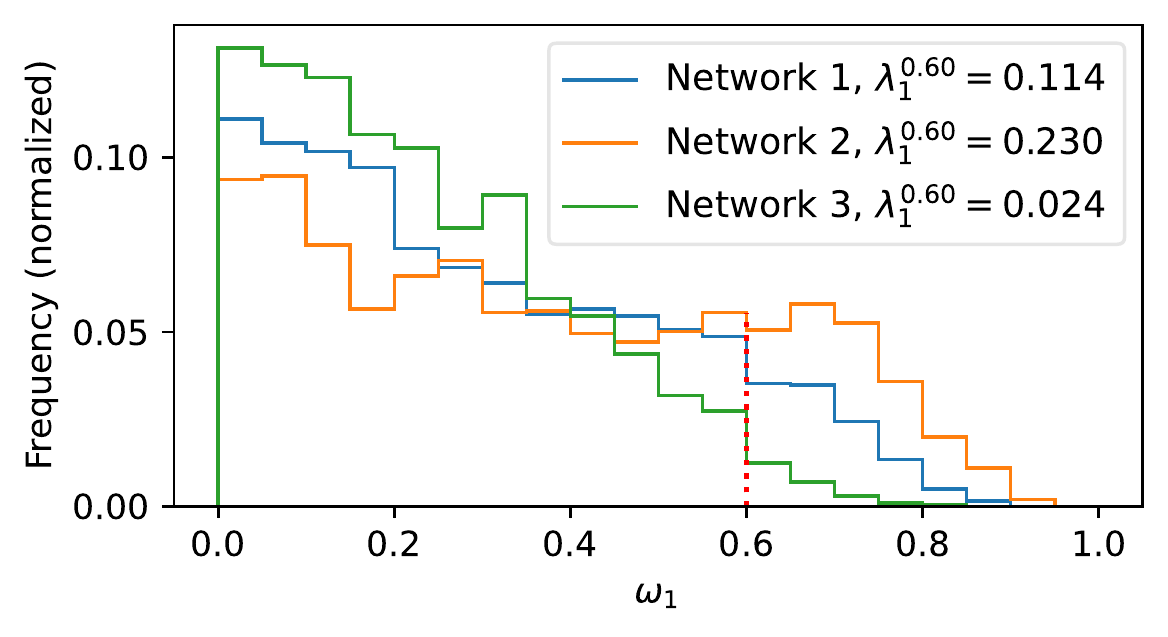}& \includegraphics[width=0.200\textwidth]{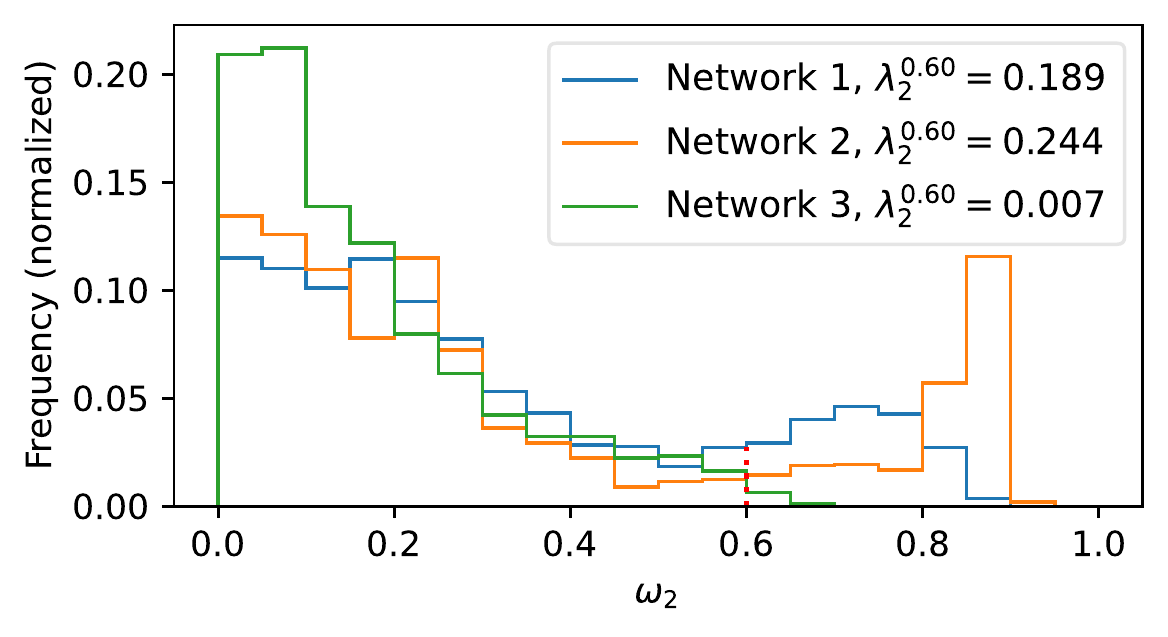} \\ \includegraphics[width=0.200\textwidth]{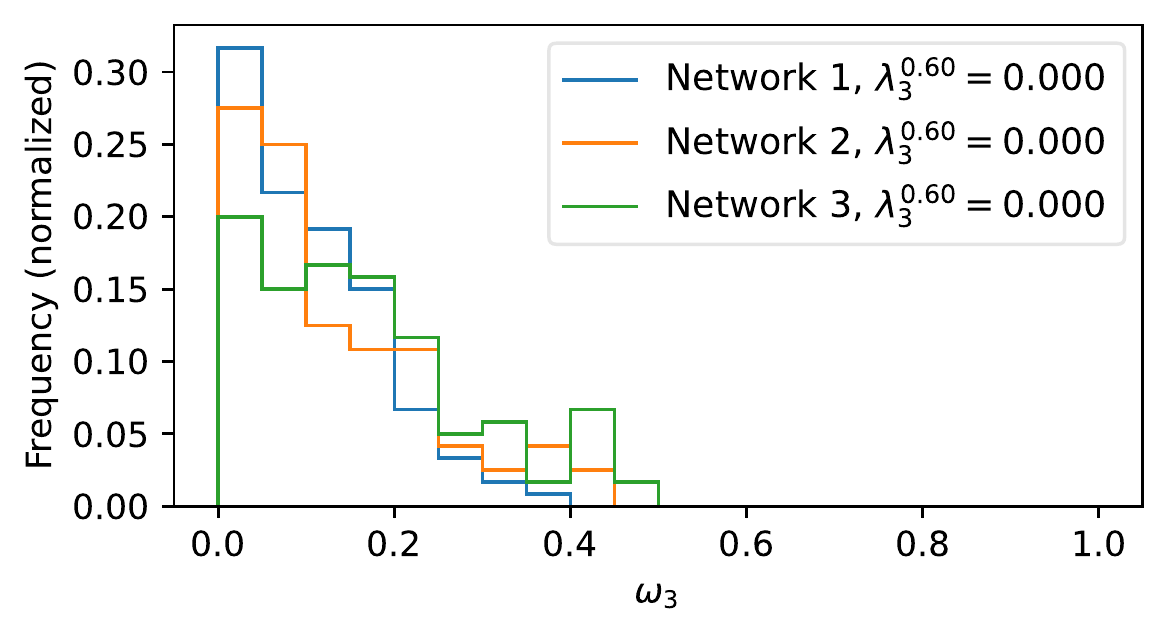}& \includegraphics[width=0.200\textwidth]{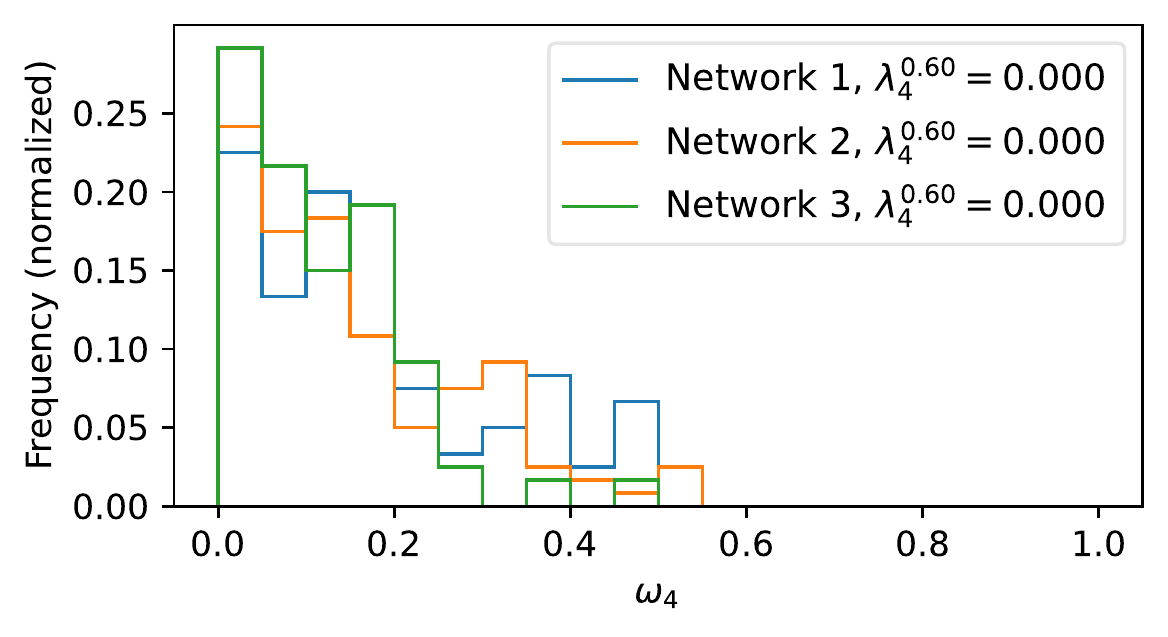} & \includegraphics[width=0.200\textwidth]{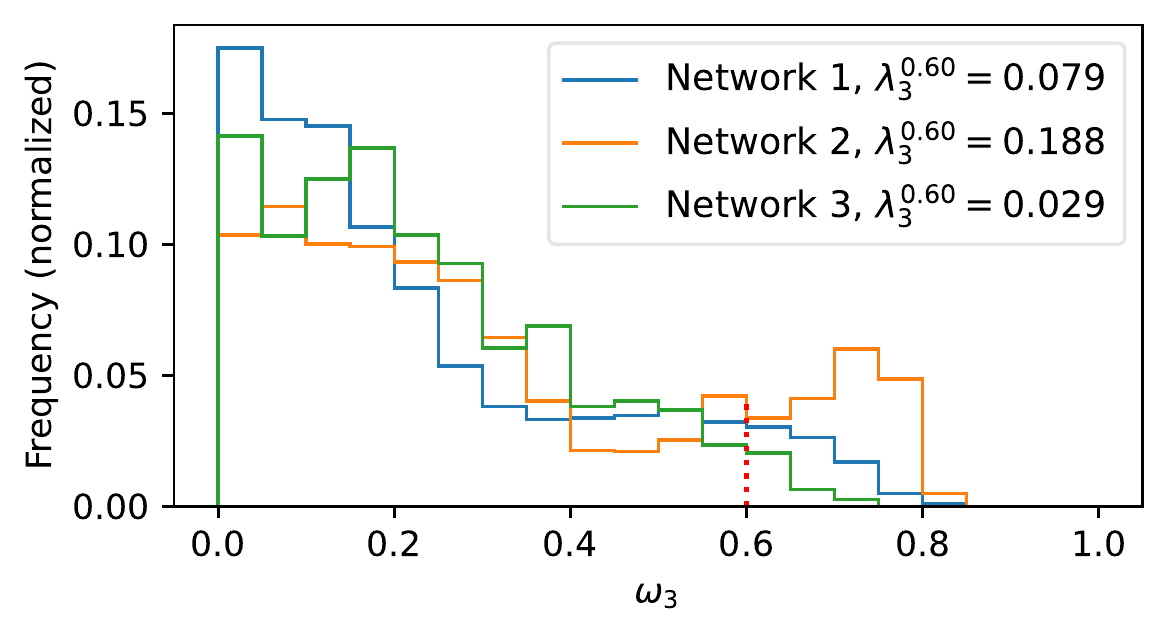}& \includegraphics[width=0.200\textwidth]{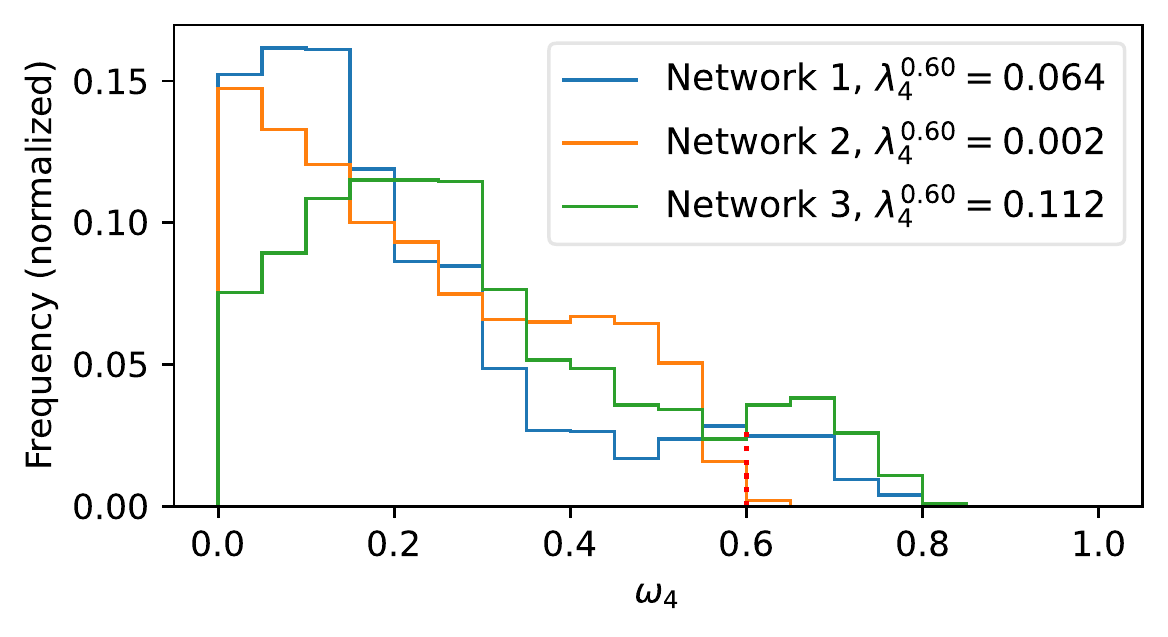} \\ \includegraphics[width=0.200\textwidth]{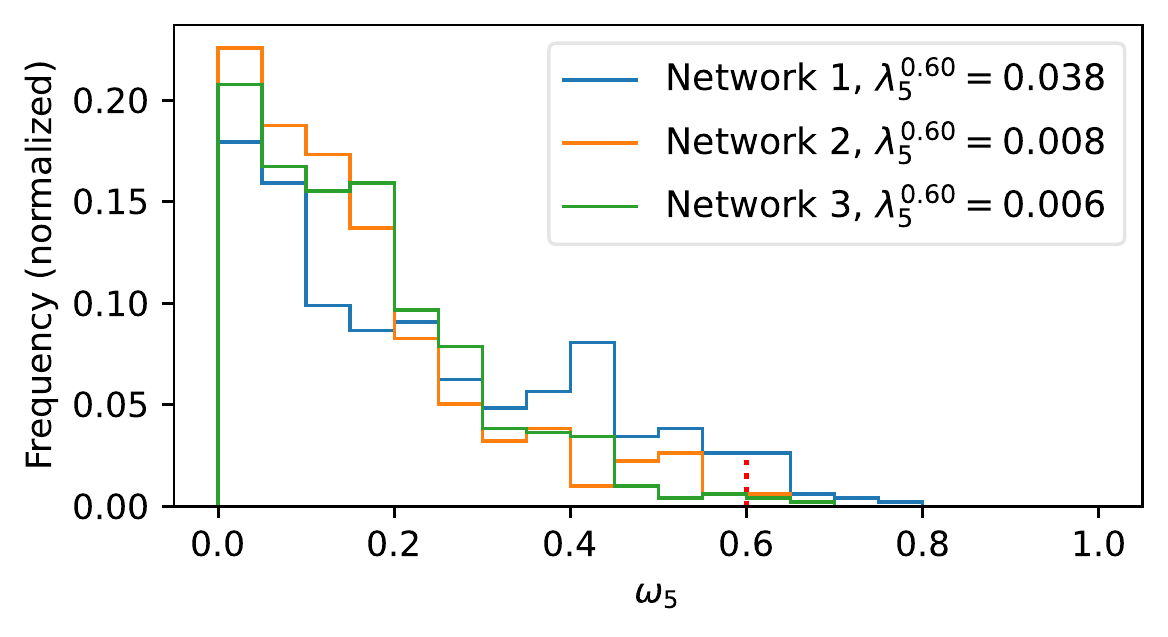}& \includegraphics[width=0.200\textwidth]{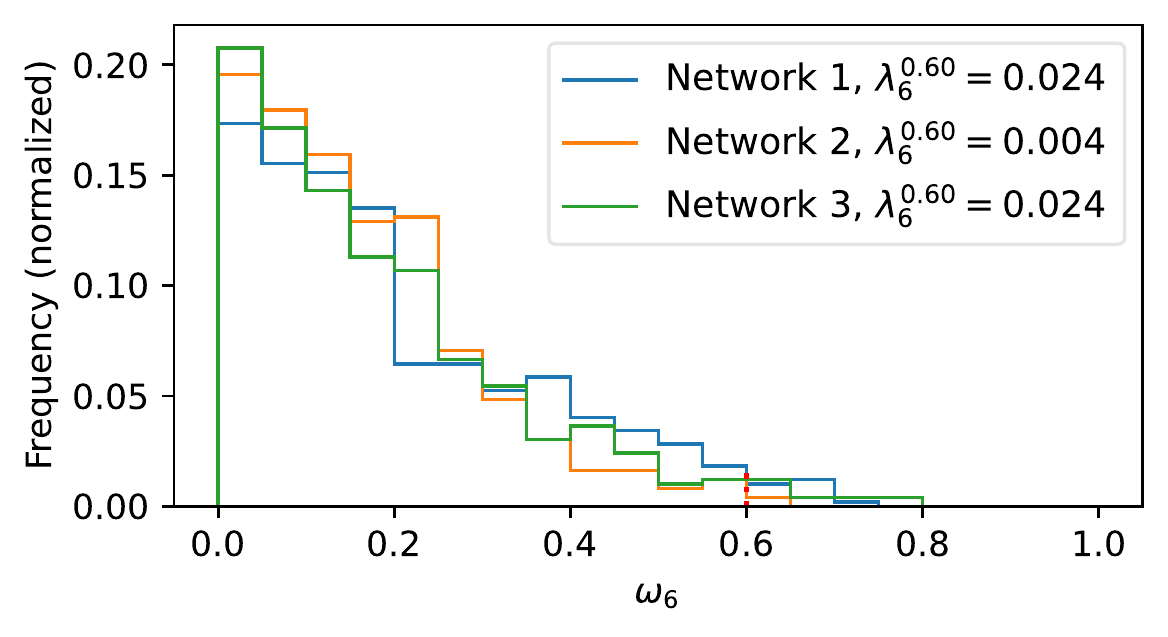} & \includegraphics[width=0.200\textwidth]{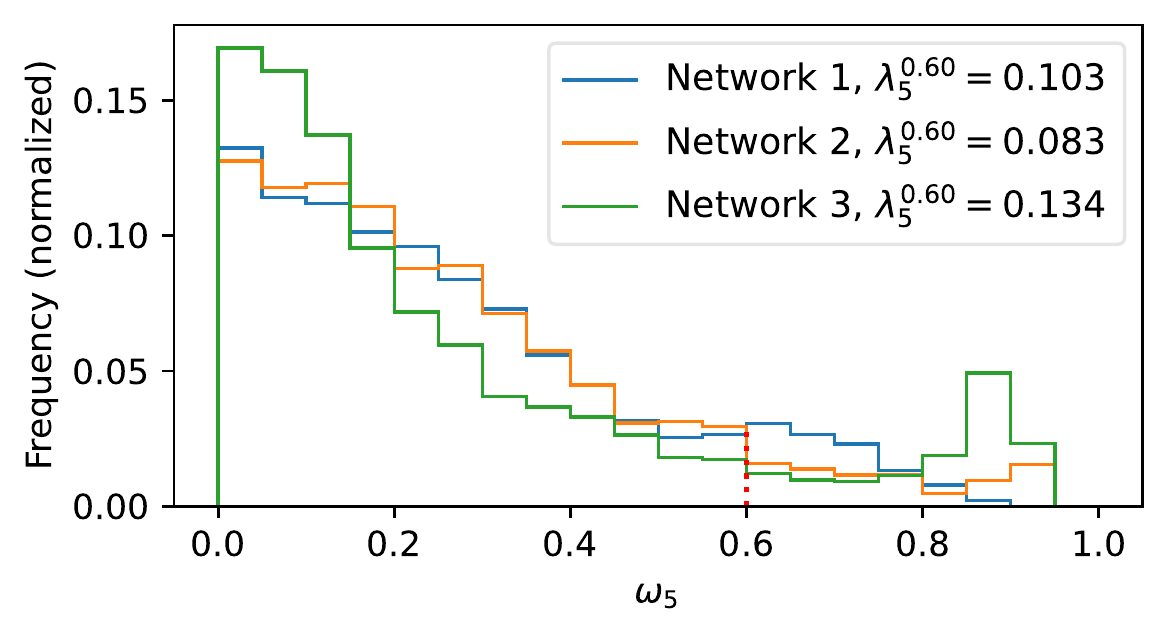}& \includegraphics[width=0.200\textwidth]{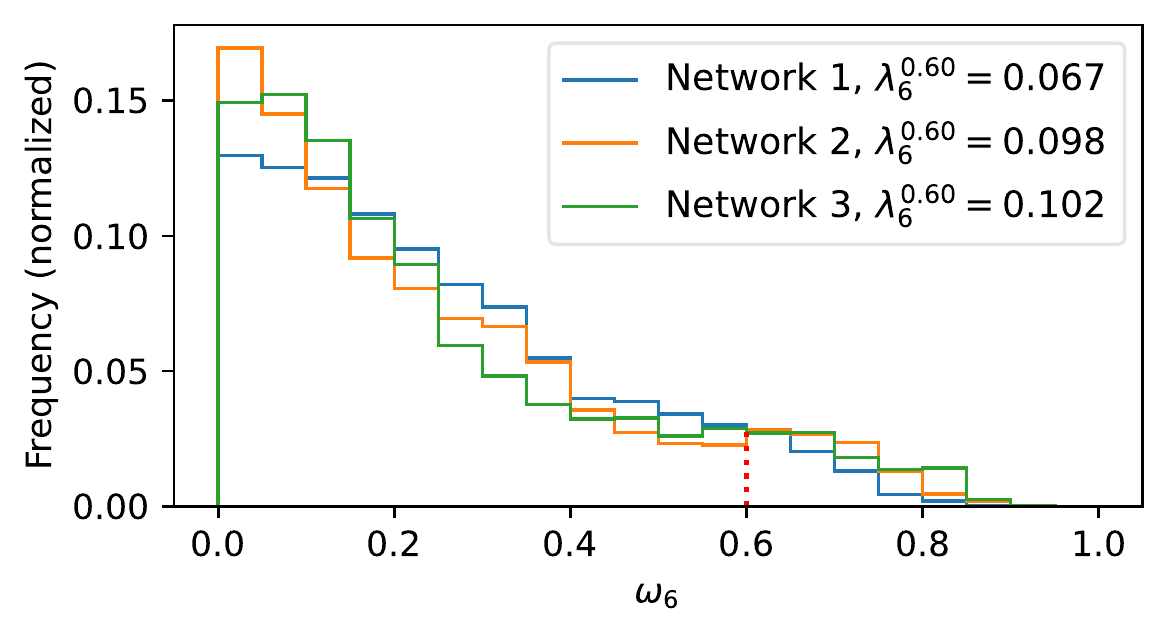} \\ \includegraphics[width=0.200\textwidth]{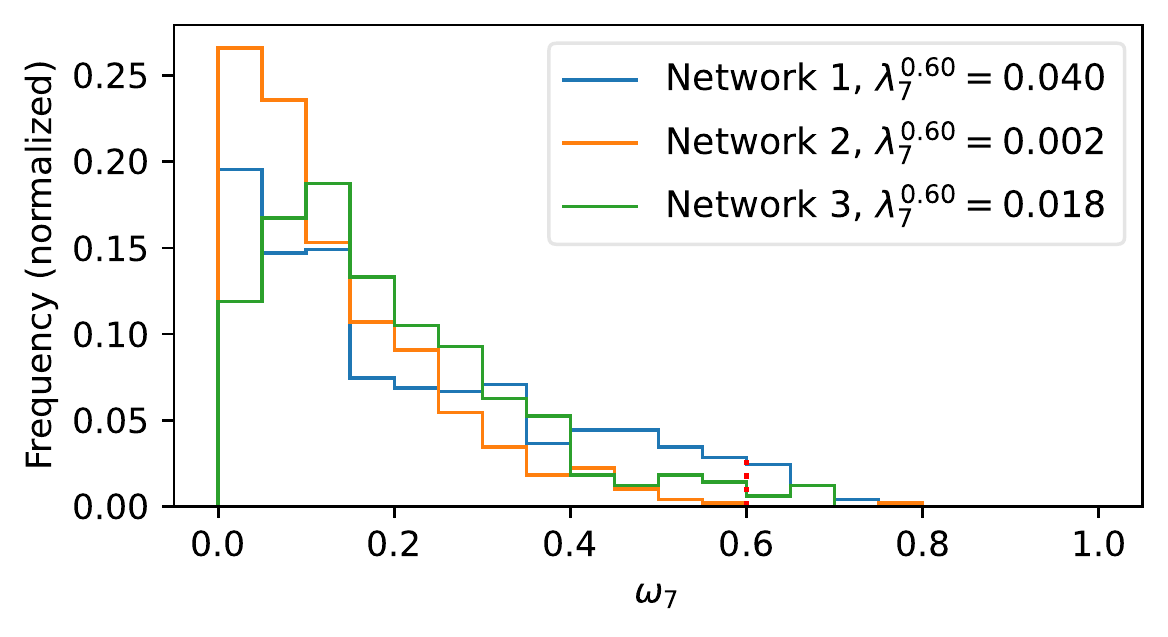}& \includegraphics[width=0.200\textwidth]{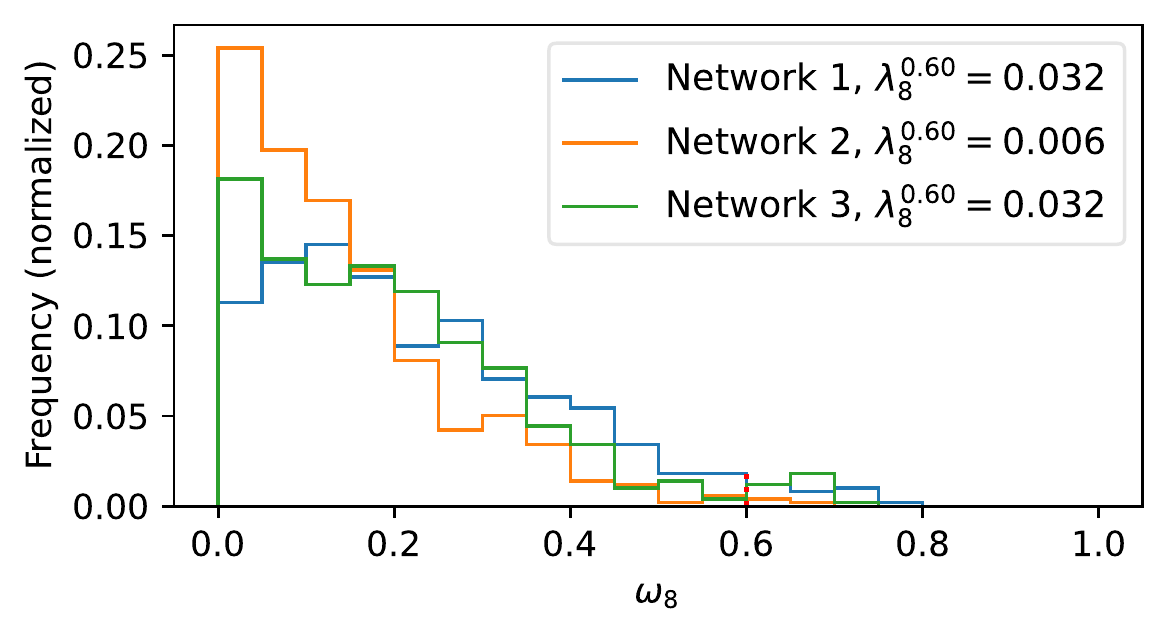} & \includegraphics[width=0.200\textwidth]{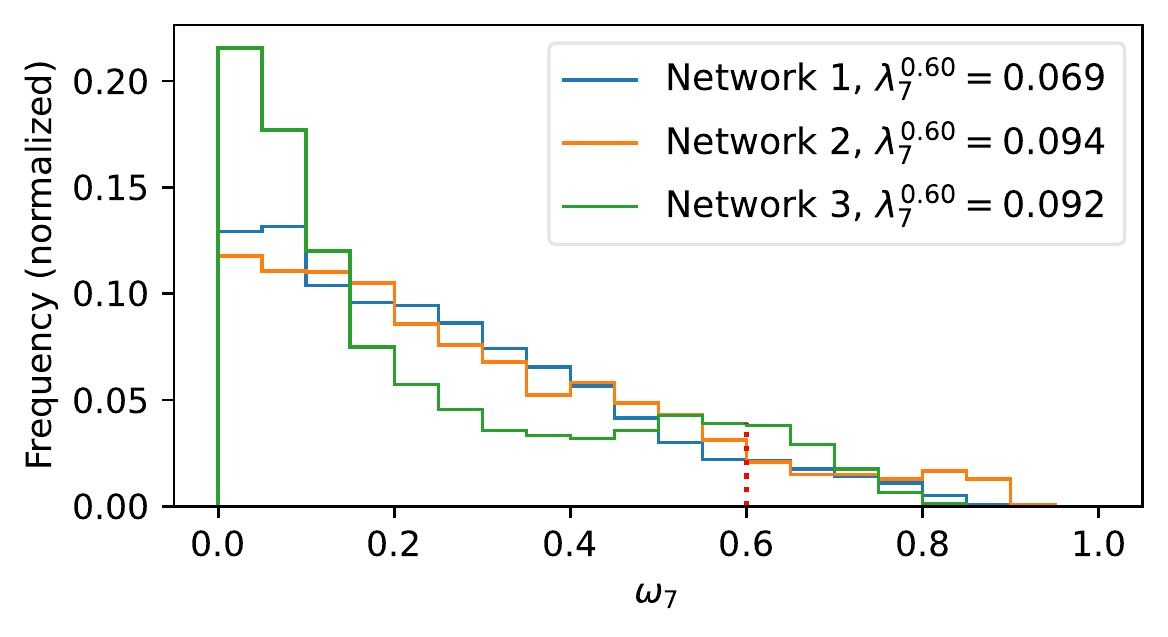}& \includegraphics[width=0.200\textwidth]{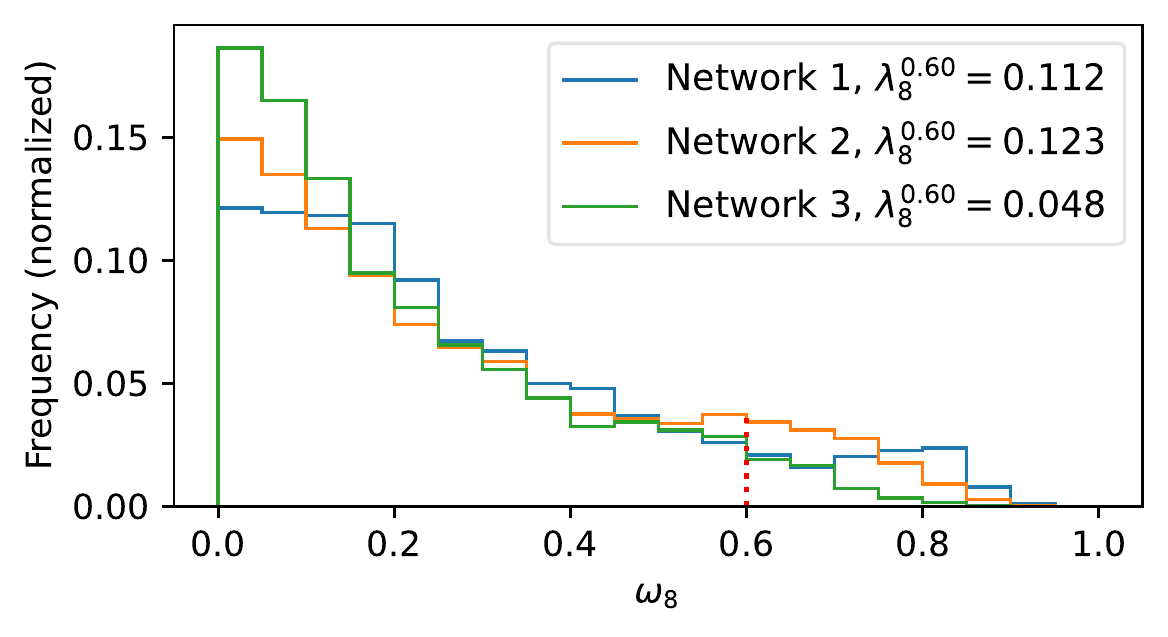} \\ \includegraphics[width=0.200\textwidth]{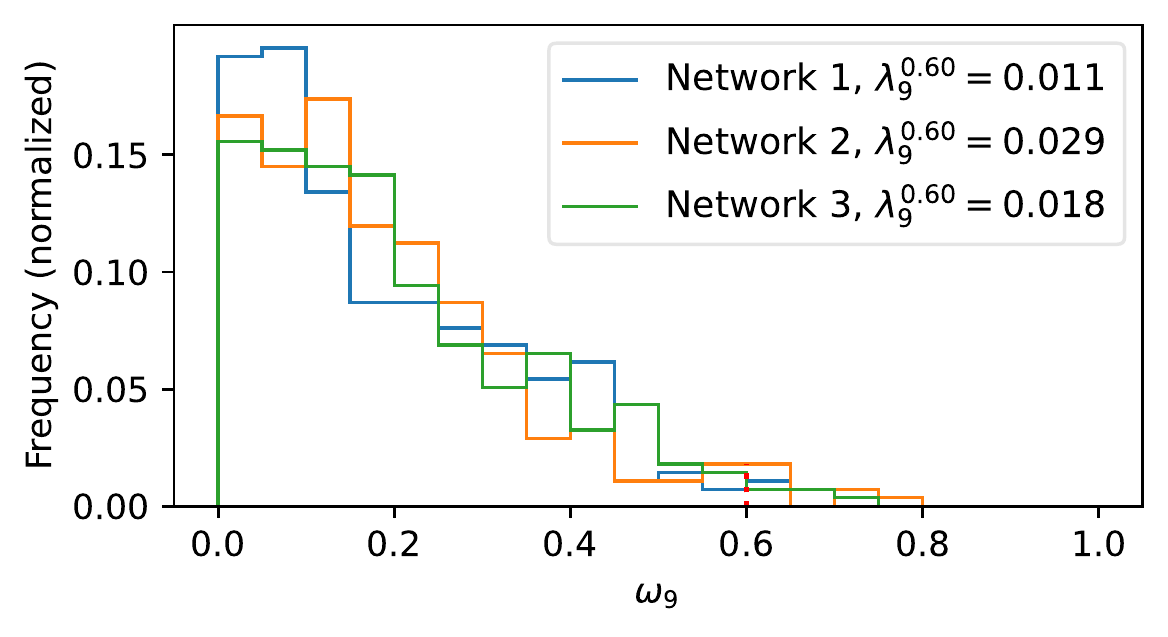}& \includegraphics[width=0.200\textwidth]{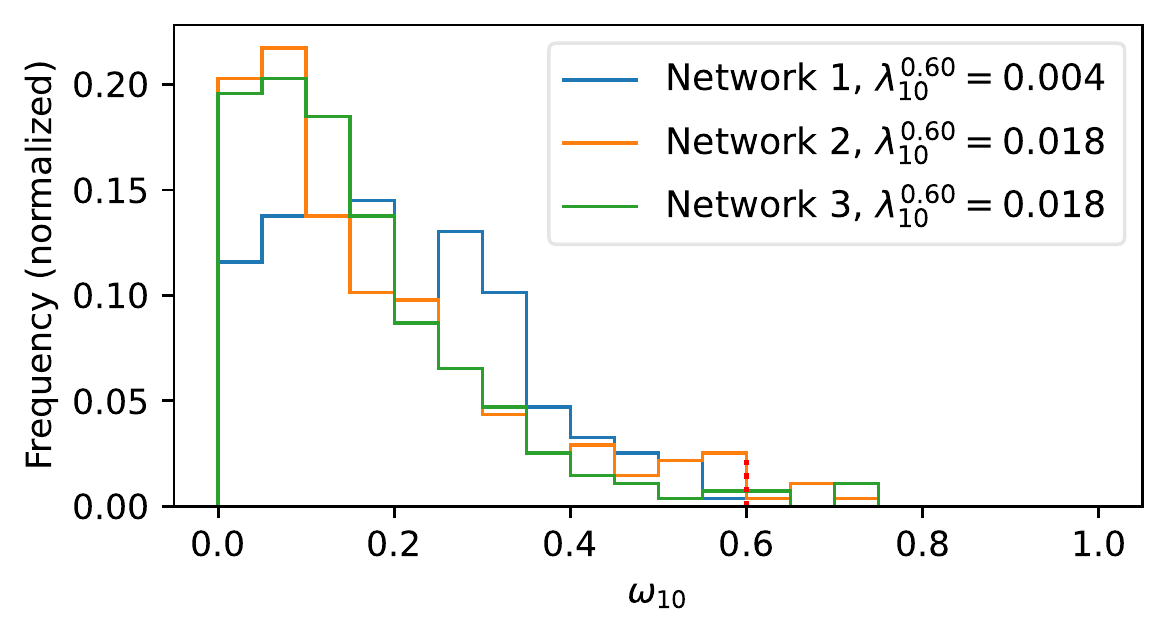} & \includegraphics[width=0.200\textwidth]{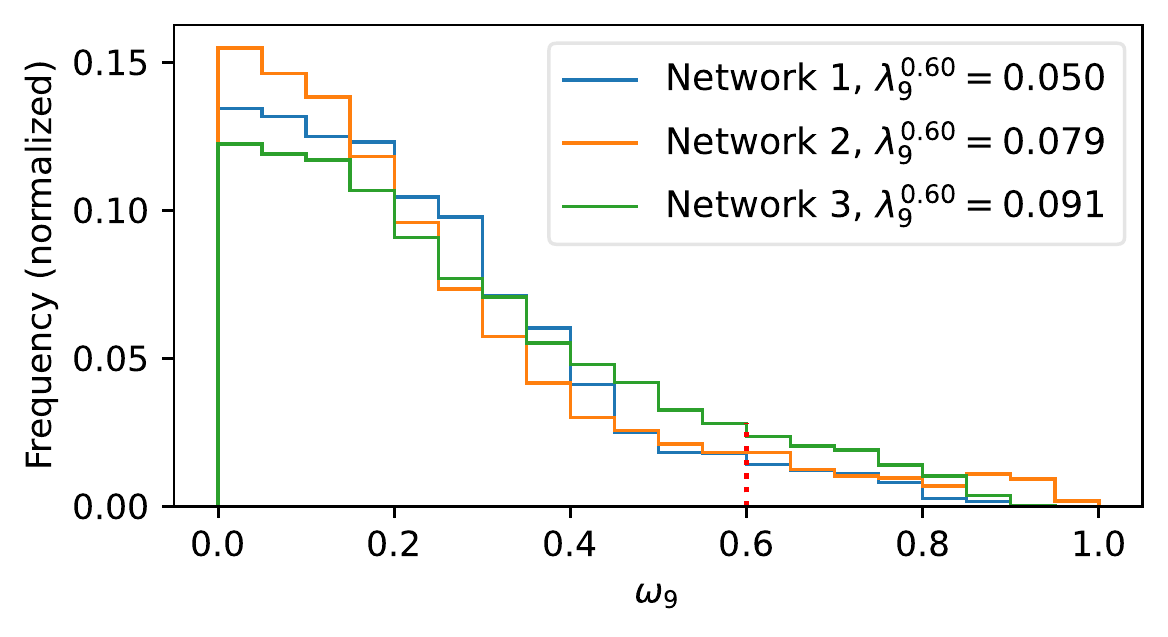}& \includegraphics[width=0.200\textwidth]{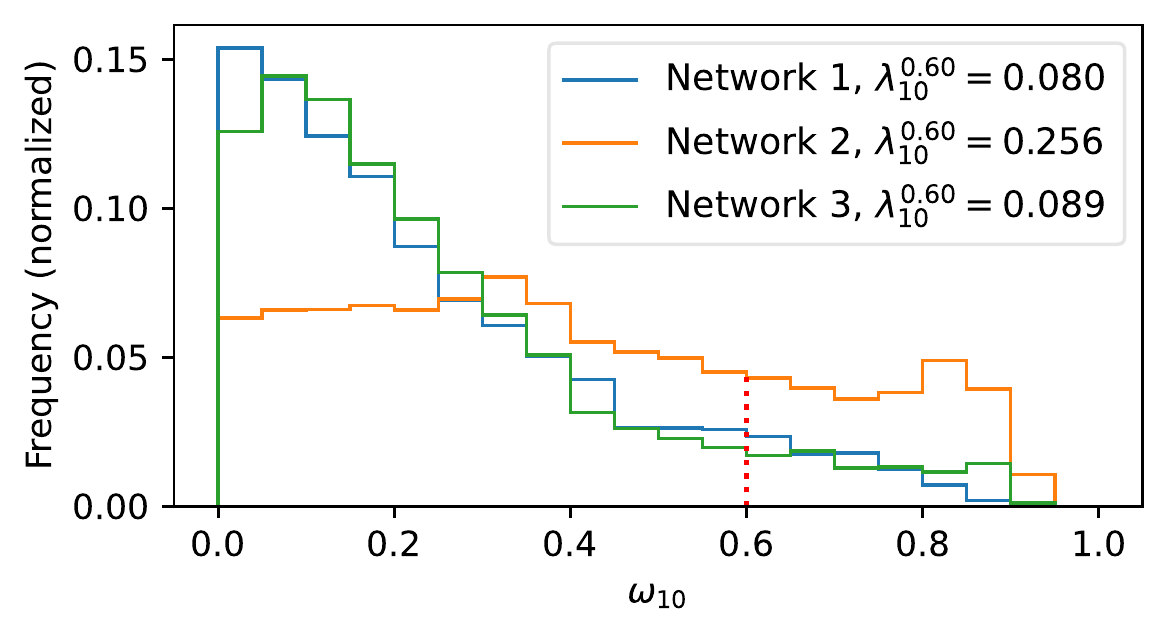} \\ \includegraphics[width=0.200\textwidth]{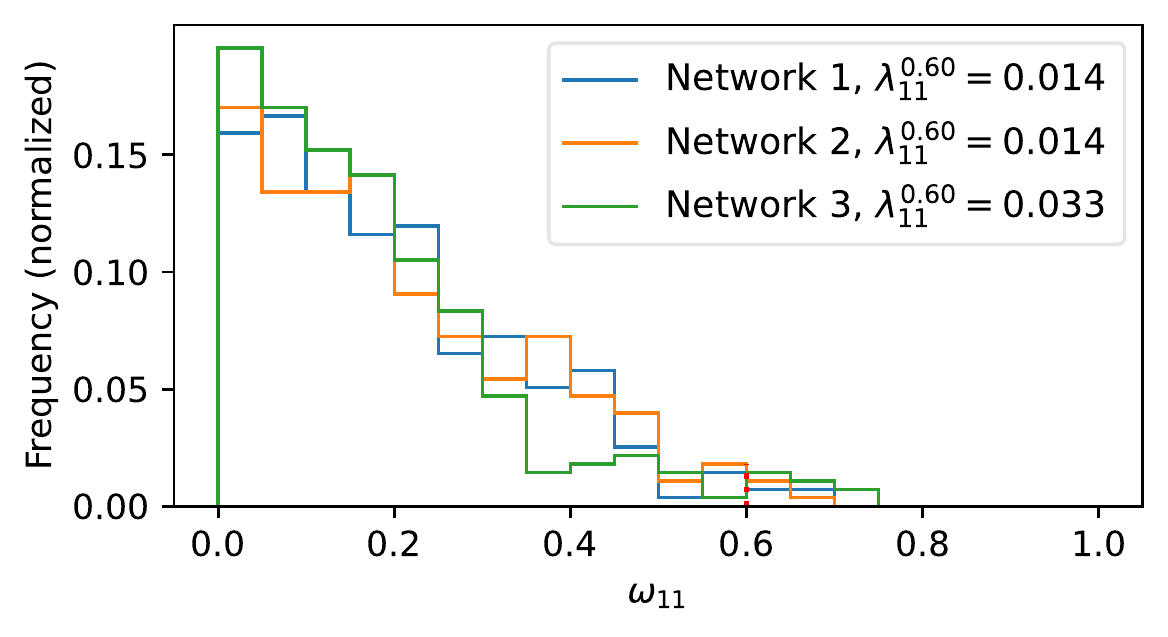}& \includegraphics[width=0.200\textwidth]{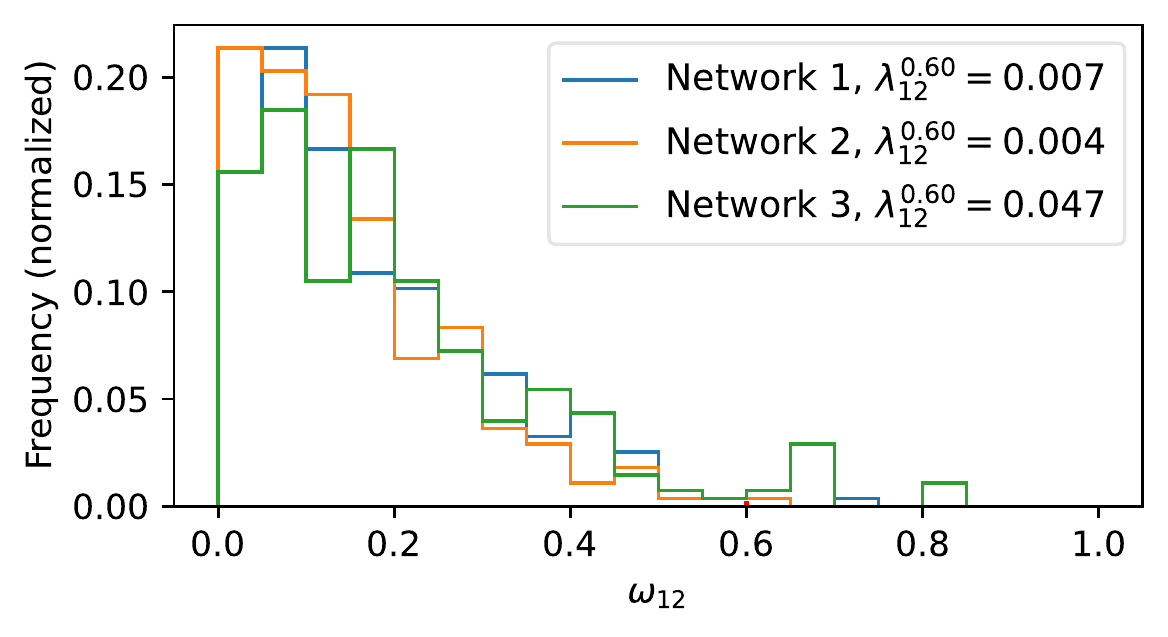} & \includegraphics[width=0.200\textwidth]{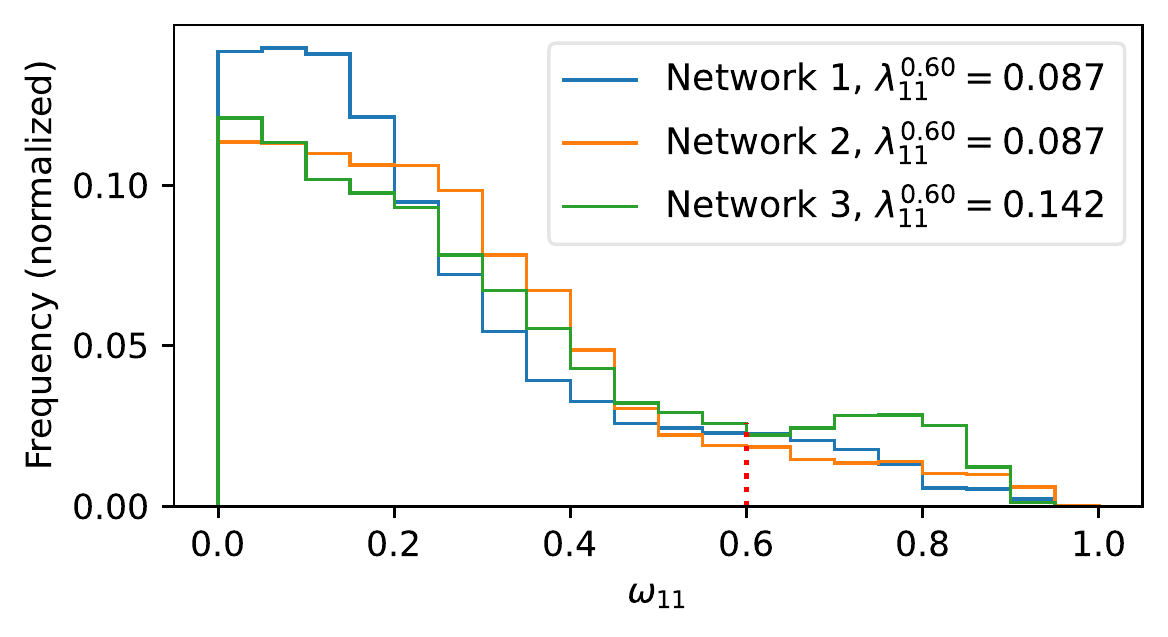}& \includegraphics[width=0.200\textwidth]{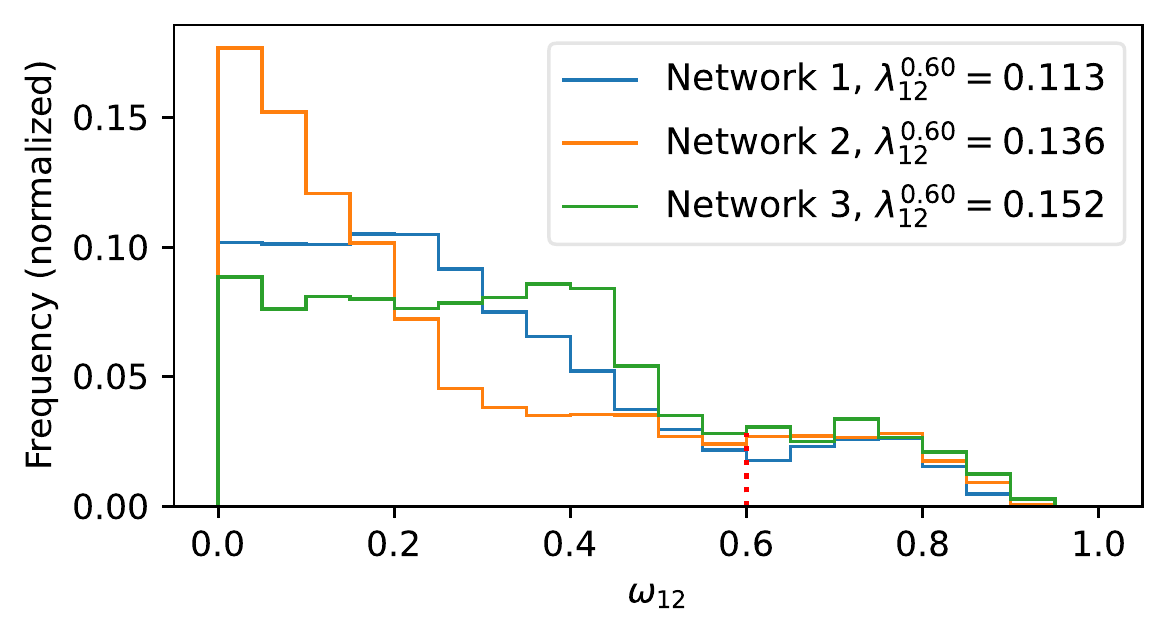} \\ \includegraphics[width=0.200\textwidth]{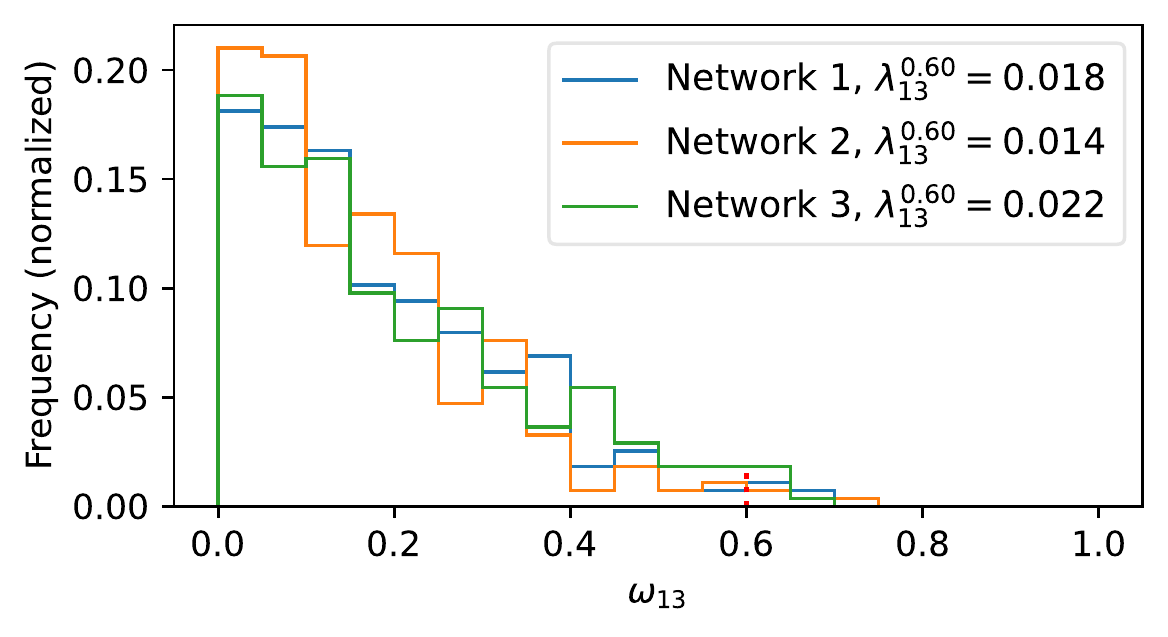}& \includegraphics[width=0.200\textwidth]{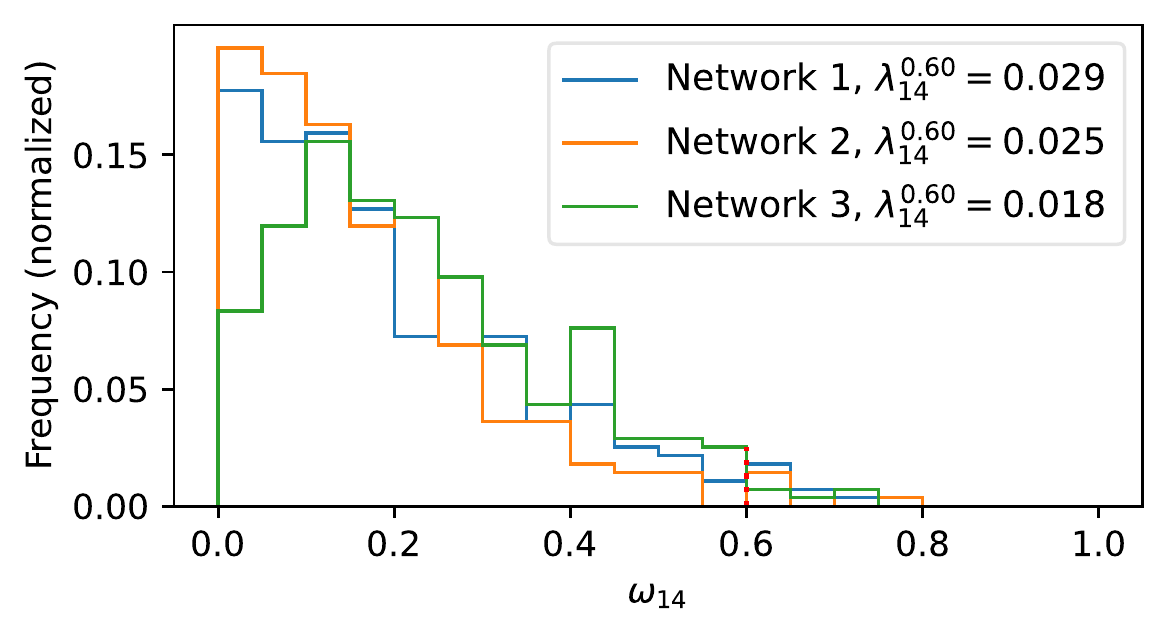} & \includegraphics[width=0.200\textwidth]{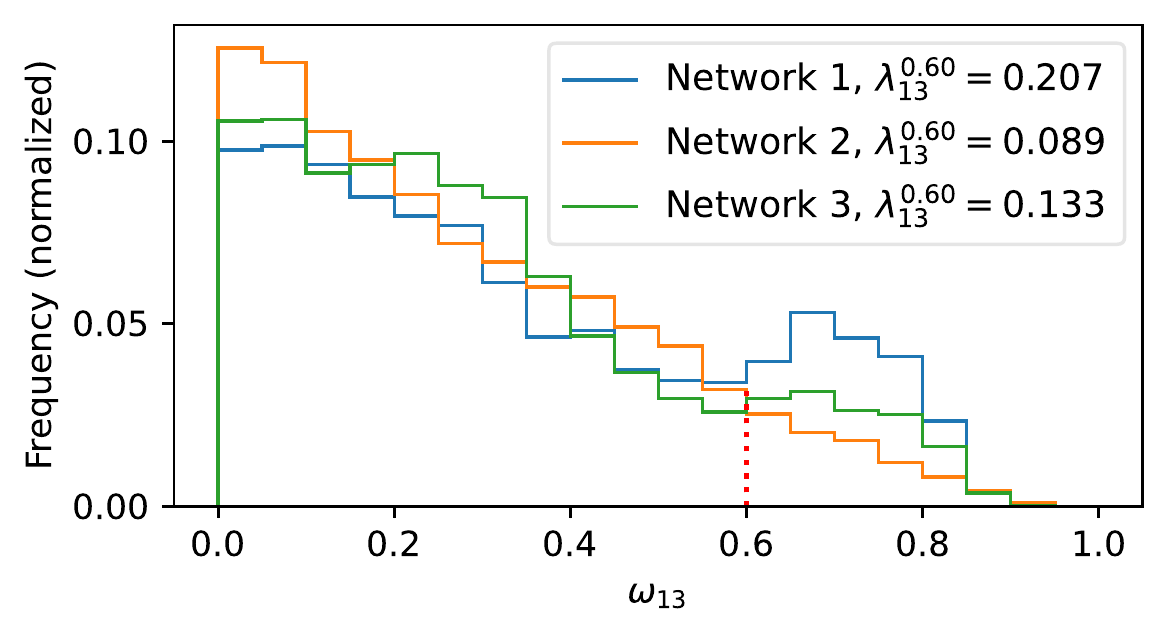}& \includegraphics[width=0.200\textwidth]{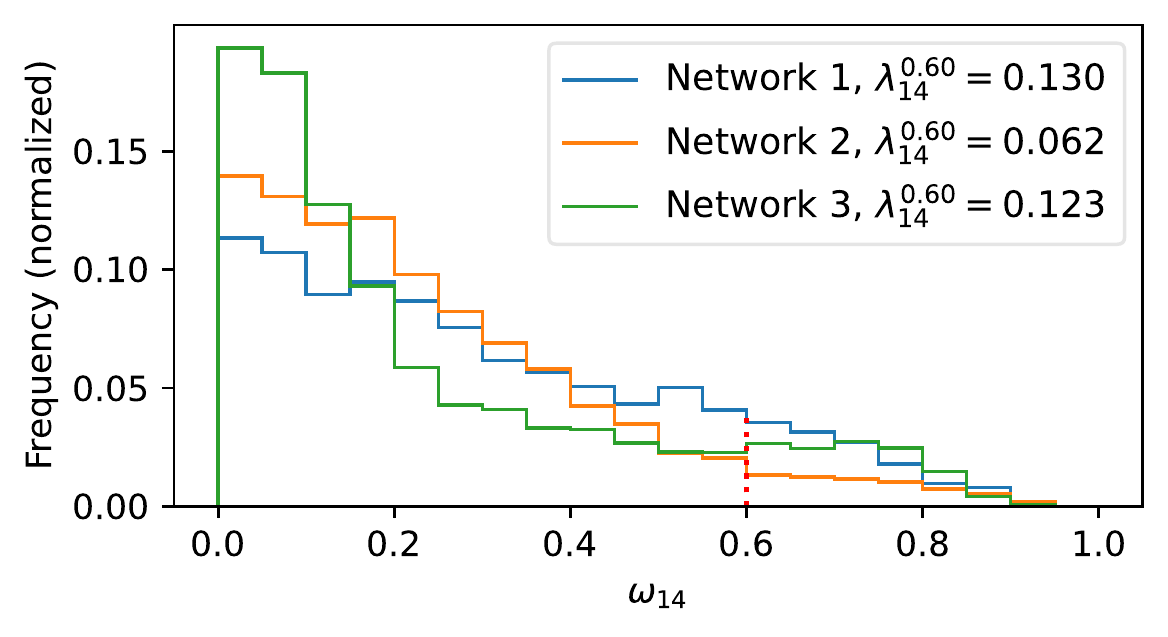} \\ \includegraphics[width=0.200\textwidth]{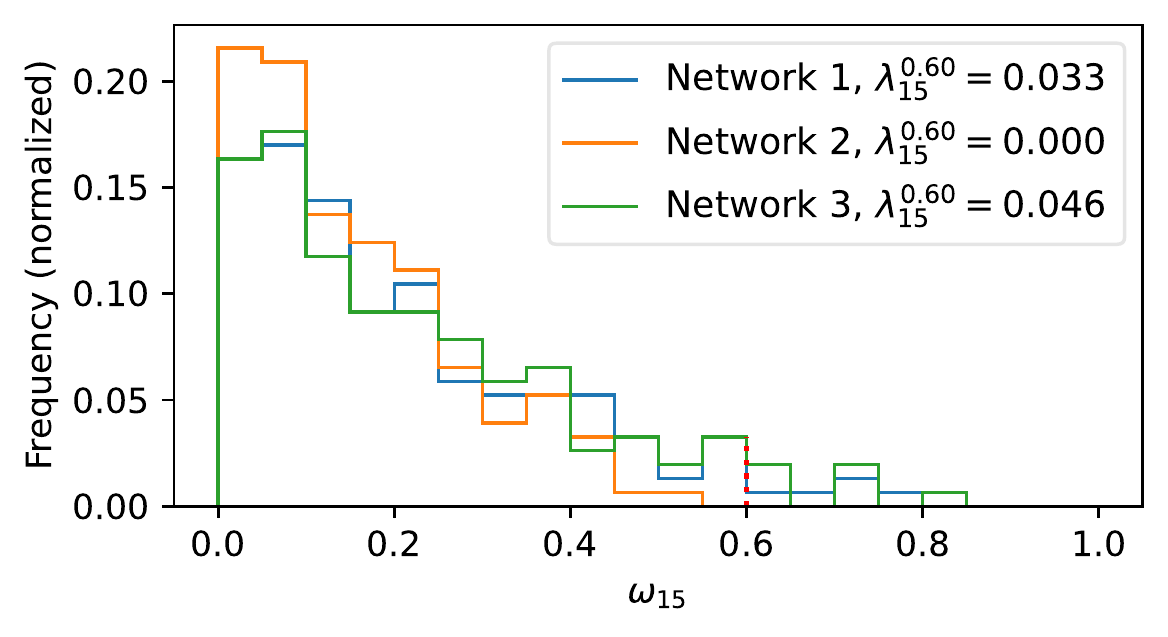}& \includegraphics[width=0.200\textwidth]{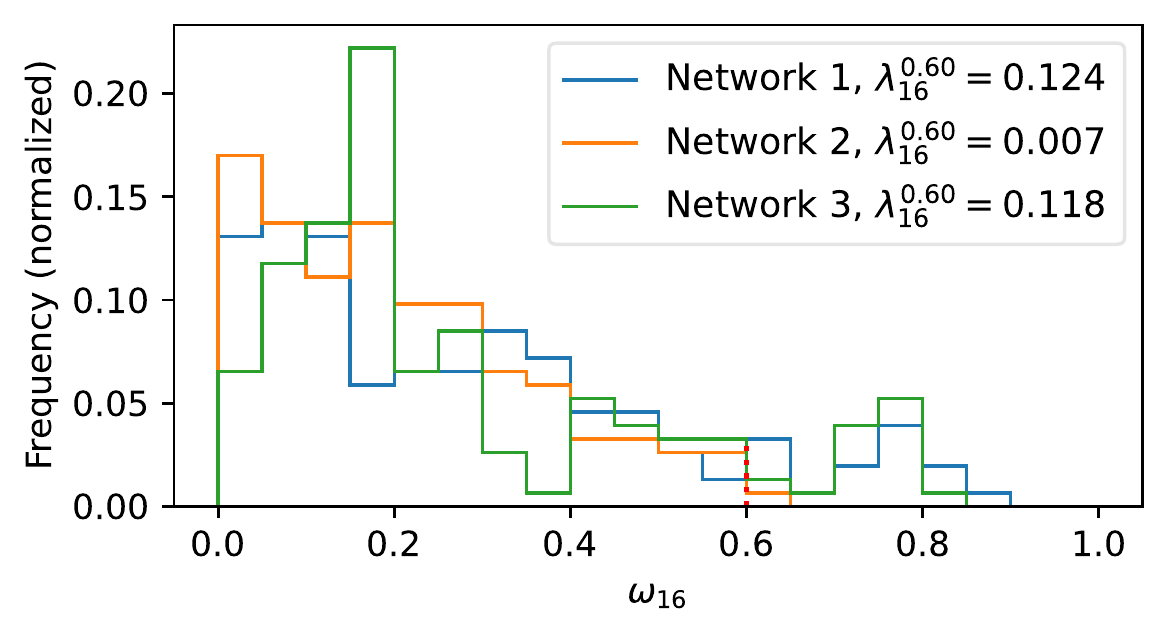} & \includegraphics[width=0.200\textwidth]{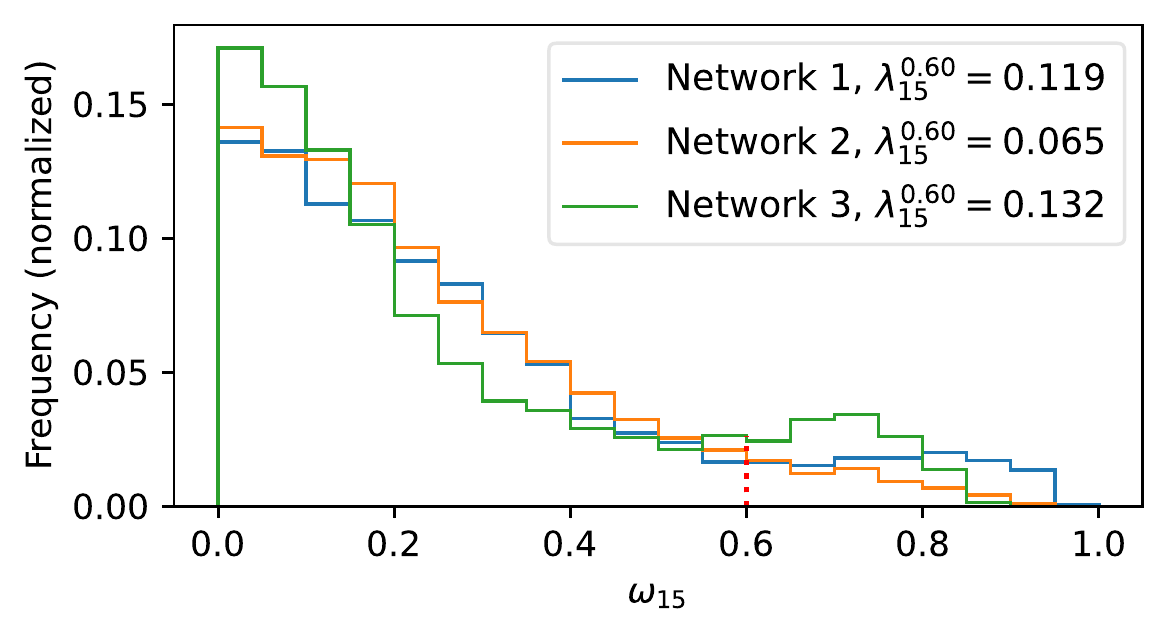}& \includegraphics[width=0.200\textwidth]{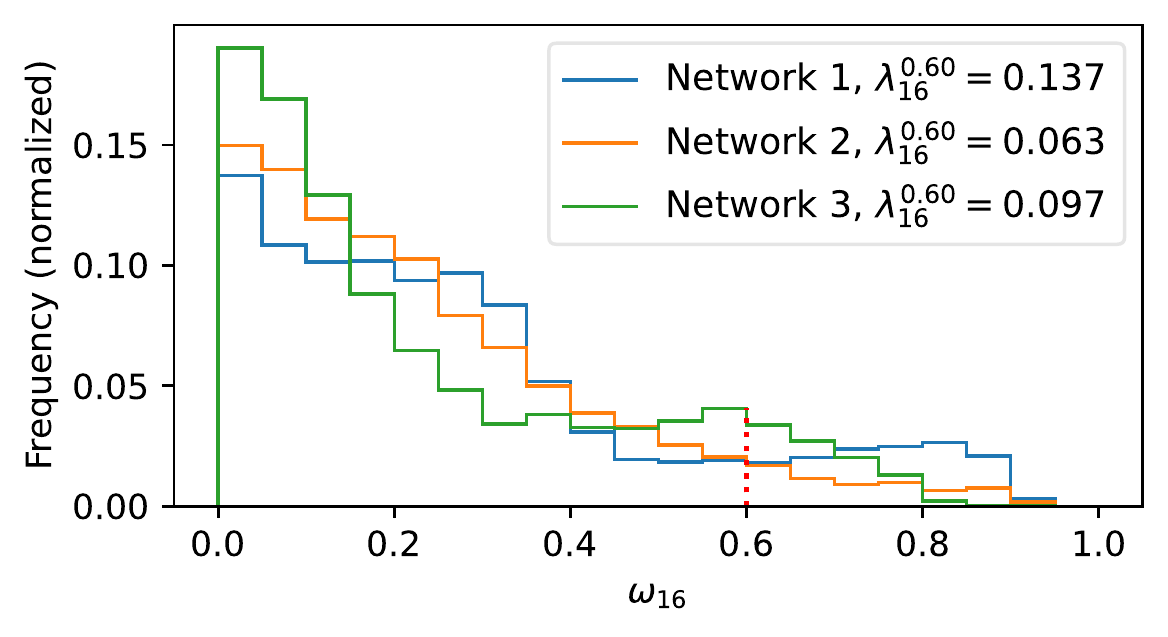} \\ \includegraphics[width=0.200\textwidth]{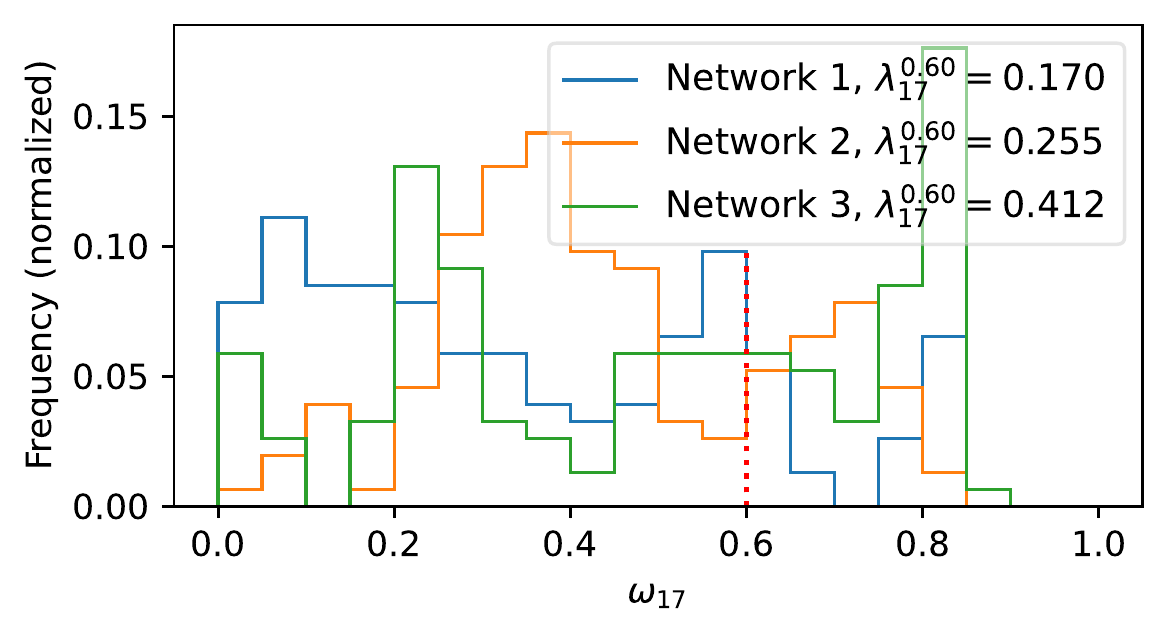}& & \includegraphics[width=0.200\textwidth]{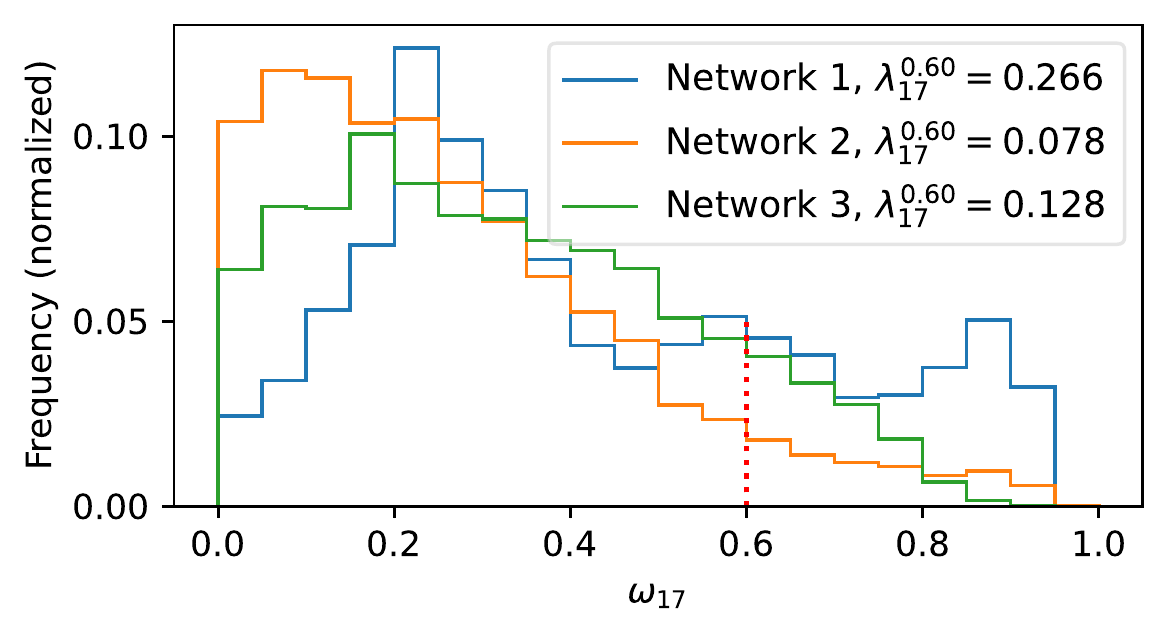}& \\
 
 \bottomrule
 
\end{tabular}
\caption{Comparison of $\omega_l$ distribution and $\lambda_l^{0.6}$ for the $17$ convolution kernels between the optimized ResNet50 (left portion) and the standardized ResNet50 (right portion). For each plot, results from three independently trained networks are presented.}
\label{tab:redundancy}
\end{table*}
 
\section{Problem Formulation}\label{sec:Problem Formulation}
 
Given the ResNet \citep{he2016deep} architectures, or other similar CNN architectures, what the network performs can be interpreted as a series of matched filtering \citep{woodward2014probability,turin1960introduction} with non-linear transformation, yet different from the classical matched filtering, because the filtering kernels are learnt during training rather than manually predefined \citep{farag2022matched}. Therefore, during training, the key functionality of the trainable convolution kernels in the convolution layers, also termed as weights, is to identify important features of the input and learn to extract useful information from the input. Let $\bm{W}_l\in\mathbb{R}^{F_{2}^{l}\times F_{1}^{l} \times K_{1}^{l} \times K_{2}^{l}}$, be a set convolution kernel at the $l$ layer, where $F_{2}^{l}$, $F_{1}^{l}$, $K_{1}^{l}$, $K_{2}^{l}\in\mathbb{Z}^+$ are kernel dimensions. $\bm{W}_l$ can be treated as a composition of $F_{2}^{l}$ independent matched filters of dimension ($F_{1}^{l} \times K_{1}^{l} \times K_{2}^{l}$), and it takes input $\bm{X}\in\mathbb{R}^{F_{1}^{l} \times W \times H}$, and produce an output $\bm{Y}\in\mathbb{R}^{F_{2}^{l} \times W \times H}$, where $W$, $H\in\mathbb{Z}^+$. For simplicity of notation, the batch dimension is neglected and we assume proper padding is performed on the input before convolution to maintain its dimension at output. Hence, for a network of $d$ convolution layers, the total number of trainable parameters in the convolution layers are:
\begin{equation}
    \bm{N}_{1:d} = \sum_{i}^d F_{2}^{i}\times F_{1}^{i} \times K_{1}^{i} \times K_{2}^{i} + F_{2}^{i}
\end{equation}
where the additional term $F_{2}^{i}$ comes from the bias in each convolution layer.
\par
Usually, as the network goes deeper ($l$ grows larger), $F_{2}^{l}$ and $F_{1}^{l}$ also increases significantly for the increased capability to identify and extract more abstract information. In ResNet50, as shown in Table \ref{tab:structure}, the first convolution kernel has a dimension of $64\times 3\times 7\times 7$, and the last convolution kernel has a dimension of $512\times 512\times 3\times 3$, as $F_{2}^{l}$ and $F_{1}^{l}$ doubles several times in some intermediate layers in the network. Although the increased convolution kernel dimension can boost the network's information processing capability, it inevitably increases the computational and memory burden at the same time. While the standardized ResNet architecture provides several predefined networks of different depths, the setting of doubling $F_{1}^{l}$ and $F_{2}^{l}$ at certain depths remains the same. The assumption that we start with $F_2^1=64$, and double $F_1^l$ and $F_2^l$ several times as the network goes deeper is worth reexamination. Through this paper, we demonstrate that we can achieve comparable network performance with a much lighter network structure (fewer trainable parameters) as shown in Table \ref{tab:structure} in the column optimized ResNet50. The key is to find the appropriate values for $F_{1}^{l}$ and $F_{2}^{l}$ for each layer while maintaining network performance.
 
\section{Solution}\label{sec:Solution}
For the first convolution layer of dimension $64\times 1\times 7\times 7$ ($64$ sets of kernels of dimension $1\times 7\times 7$), one can visually inspect the shape of the learnt kernels, and identify different kinds of low-level features used by the CNN. However, visual inspection of kernel shapes beyond the first convolution layer is impractical, both due to the abstract nature of kernels, and the growing dimension of kernels. 
For a CNN, there might be tens and even hundreds of convolution layers, making the optimization on $F_{2}^{l}$ and $F_{1}^{l}$ for each layer via manual tuning or random search impractical. Therefore, we propose the quantifiable perception-based ResNet structure optimization method as an efficient and visualizable solution. The essence of our method is the definition of the quantity termed convolutional kernel redundancy measure (CKRM), through which we can determine if there is a possibility to reduce $F_{2}^{l}$ and $F_{1}^{l}$ for the convolution layer. For the rest of this section, we will present the structural similarity index measure (SSIM) based on which the CKRM is derived, and then propose the CKRM. We also present how to adjust $F_{2}^{l}$ and $F_{1}^{l}$ with CKRM.
 
\subsection{Structural similarity index measure}
SSIM is one of the most frequently used image similarity measures proposed by \citeauthor{wang2004image}. Given two images, $\bm{I}_1$ and $\bm{I}_2$, SSIM takes two patches from the image at the same position, such that $\bm{x}\subseteq\bm{I}_1$ and $\bm{y}\subseteq\bm{I}_2$, and compute luminance similarity (denoted as $\mathcal{L}(\bm{x}, \bm{y})$), contrast similarity (denoted as $\mathcal{C}(\bm{x}, \bm{y})$), and structure similarity (denoted as $\mathcal{S}(\bm{x}, \bm{y})$) as:
\begin{equation} \label{eq:LCS}
\begin{split}
\mathcal{L}(\bm{x}, \bm{y}) & = \frac{2\mu_x\mu_y+\epsilon}{\mu_x^2+\mu_y^2+\epsilon} \\
\mathcal{C}(\bm{x}, \bm{y}) & = \frac{2\sigma_x\sigma_y+\epsilon}{\sigma_x^2+\sigma_y^2+\epsilon} \\
\mathcal{S}(\bm{x}, \bm{y}) & = \frac{\sigma_{xy}+\epsilon}{\sigma_x\sigma_y+\epsilon} \\
\end{split}
\end{equation}
where $\mu_x$ and $\mu_y$ stand for the mean of elements in $\bm{x}$ and $\bm{y}$, respectively; $\sigma_x$ and $\sigma_y$ stand for the standard deviation of elements in $\bm{x}$ and $\bm{y}$, respectively; $\sigma_{xy}$ is the covariance between $\bm{x}$ and $\bm{y}$; $\epsilon$ is a small constant value to avoid division by zero. SSIM between $\bm{I}_1$ and $\bm{I}_2$, denoted as $\Phi$, is calculated as \citep{wang2003multiscale}:
\begin{equation}
    \Phi(\bm{I}_1, \bm{I}_2) = \frac{1}{n}\sum_{i=1}^n \phi(\bm{x}_i, \bm{y}_i)
    \label{eq:Phi}
\end{equation}
where the patch-wise SSIM, denoted as $\phi$, is calculated as:
\begin{equation}
    \phi(\bm{x}, \bm{y}) = \mathcal{L}(\bm{x}, \bm{y}) \cdot \mathcal{C}(\bm{x}, \bm{y}) \cdot \mathcal{S}(\bm{x}, \bm{y})
    \label{eq:phi}
\end{equation}
where $\bm{x}$ and $\bm{y}$ can be any different sets of patches from $\bm{I}_1$ and $\bm{I}_2$, respectively; $n\in\mathbb{Z}^+$.
\par
$\Phi(\bm{I}_1, \bm{I}_2)$ has a range of $[-1,1]$ where the closer the value is to $0$, the more different the images are. $\Phi(\bm{I}_1, \bm{I}_2)=	\pm 1$ when $\bm{I}_1 \pm \bm{I}_2$.
 
\subsection{Convolution kernel redundancy measure}
For a convolution layer with weights $\bm{W}_l$ of dimension $F_{2}^{l}\times F_{1}^{l} \times K_{1}^{l} \times K_{2}^{l}$, it is composed of a set of $F_{2}^{l}$ different kernels, $\{\bm{W}_l^1,\bm{W}_l^2,...,\bm{W}_l^{F_{2}^{l}}\}$, each of size $F_{1}^{l} \times K_{1}^{l} \times K_{2}^{l}$. $F_{1}^{l}$ is the number of channels in the input. Let $\bm{w}_l^{i,j}\in\mathbb{R}^{K_{1}^{l} \times K_{2}^{l}}$ be the $(i,j)$ slice upon the first two dimensions of the weights $\bm{W}_l$. We can use the SSIM to quantify the similarity between $\bm{w}_l^{p,q}$ and $\bm{w}_l^{i,j}$, $p$, $q$, $i$, $j\in\mathbb{Z}^+$. A modification of Equation (\ref{eq:phi}) needs to be made to better quantify the similarity between convolution kernels. Following \citeauthor{wang2003multiscale}, we use the following equation to calculate the similarity between two kennels of size $K_{1}^{l} \times K_{2}^{l}$:
\begin{equation}
    \psi_{\alpha, \beta, \gamma}(\bm{x}, \bm{y}) = \abs{\mathcal{L}(\bm{x}, \bm{y})}^\alpha \cdot \left[\mathcal{C}(\bm{x}, \bm{y})\right]^\beta \cdot \abs{\mathcal{S}(\bm{x}, \bm{y})}^\gamma
    \label{eq:psi}
\end{equation}
where $\alpha$, $\beta$, and $\gamma$ are positive weights. Comparing Equations (\ref{eq:phi}) and (\ref{eq:psi}), we can see that the following equality holds:
\begin{equation}
    \psi_{1, 1, 1}(\bm{x}, \bm{y}) = \abs{\phi(\bm{x}, \bm{y})}.
\end{equation}
In another word, $\psi_{\alpha, \beta, \gamma}(\bm{x}, \bm{y})$ is a generalization of $\phi(\bm{x}, \bm{y})$ that allows us to allocate $\mathcal{L}(\bm{x}, \bm{y})$, $\mathcal{C}(\bm{x}, \bm{y})$, and $\mathcal{S}(\bm{x}, \bm{y})$ with different weight during computation. In this paper, we choose $\alpha=0.1$, $\beta=1$, and $\gamma=1$. Therefore, we define the measurement of kernel similarity between the $i$ and $j$ kernels, $\bm{W}_l^i$ and $\bm{W}_l^j$, in convolution layer $l$ as:
\begin{equation}
    \Psi_l(i,j) = \frac{1}{F_{1}^l}\sum_{k=0}^{F_{1}^l} \psi_{0.1, 1, 1}(\bm{w}_l^{i,k}, \bm{w}_l^{j,k}).
    \label{eq:Psi}
\end{equation}
From Equation (\ref{eq:Psi}), we know that $\Psi_l(i,j)\in[0,1]$, and $\Psi_l(i,j)=1$ when $\bm{W}_l^i=\pm\bm{W}_l^j$. The larger the $\Psi_l(i,j)$, the more similarities can be found between $\bm{W}_l^i$ and $\bm{W}_l^j$.
\par
The main reason why Equation (\ref{eq:Psi}) is better fitted for measuring kernel similarity than Equation (\ref{eq:Phi}) lies in the fact that, for a CNN, the shape of a kernel matters more than the precise values of a kernel. That is to say, for $\bm{W}_l^i$ and $\bm{W}_l^j$, if $\bm{W}_l^i\approx \bm{W}_l^j+a$, where $a\in\mathbb{R}$, they are extracting very similar kind of information from the input. Therefore, the defined similarity measure should be large if 
$\bm{W}_l^i\approx \bm{W}_l^j+a$. That is true for Equation (\ref{eq:Psi}) but not for Equation (\ref{eq:Phi}). This can be better visualized with Table \ref{tab:demo}. From left to right, zero mean additive white Gaussian noise with increasing variance is added. We can see that for the top row image, $\psi_{1,1,1}$ and $\psi_{0.1,1,1}$ do not exhibit major differences, as the source of the dissimilarity are mostly from contrast and structure change and not due to luminance differences. From the top row images to the bottom row images, we increase the luminance with a constant value of $0.5$, both $\psi_{1,1,1}$ and $\psi_{0.1,1,1}$ drops, with $\psi_{0.1,1,1}$ dropping less significantly, as we have tuned down the weight for luminance similarity in Equation (\ref{eq:Psi}).
\par
With $\Psi_l(i,j)$, we define the set of pairwise similarity measure for kernels in $\bm{W}_l$ as:
\begin{equation}
    \bm{\Omega}_l = \left\{\Psi_l(i,j) \: \vline \:  i<j; i,j\in\{1,2,...,F_2^l\}\right\}.
\end{equation}
The set $\bm{\Omega}_l$ contains $F_2^l(F_2^l-1)/2$ elements, where each element represents the similarity measure between a pair of kernels in $\bm{W}_l$. The convolution kernel redundancy measure (CKRM) is defined as:
\begin{equation}
    \lambda_{l}^t = \mbox{Pr}(\omega_l\gt t)
    \label{eq:lambda}
\end{equation}
where $\mbox{Pr}(\Omega_l>t)$ represents the probability that a element in $\bm{\Omega}_l$, $\omega_l$, has a value greater than the threshold $t\in[0,1]$.
\par
A small $\lambda_{l}^t$ is desirable $\forall t\in[0,1]$, as we wish that the convolution kernels can focus on input features of various forms. In the extreme case when $\lambda_{l}^t=1$,  $\forall t\in[0,1)$, all the convolution kernels are only focusing on a single feature, which is a waste of the memory and computation resource, and thus should be avoided.
\par
As $F_l^2$ can be very large, i.e., $F_l^2=512$, computing all elements in $\bm{\Omega}_l$ can be time consuming. Yet with the probabilistic nature of Equation (\ref{eq:lambda}), we can sample from $\bm{\Omega}_l$, and estimate $\lambda_{l}^t$ with a subset of $\bm{\Omega}_l$. 
 
\subsection{Network Structure Optimization with CKRM}
As the training of a CNN is a stochastic process, the resulting kernels in the convolutions layers can also vary from one training outcome to another. That is one of the reasons why we use the probabilistic notation in Equation (\ref{eq:lambda}) to define $\lambda_{l}^t$. In an empirical study, we found that despite the arbitrary shape of the learnt kernels usually varying as a result of independent training, the distribution of $\omega_l$, and the resulting $\lambda_{l}^t$ values are approximately the same. Table \ref{tab:redundancy} shows the $\omega_l$ distribution and the resulting $\lambda_{l}^{0.6}$ values. The threshold of $0.6$ is chosen as the target, and we wish to make $\lambda_{l}^{0.6}$ as close to zero as possible. From Table \ref{tab:redundancy}, we can see that for all convolution kernels, $\omega_l$ has a long-tailed positive skew distribution with some values located close to $1$. We wish to eliminate large $\omega_l$ values that are greater than the specified threshold. That is achieved by reducing the number of filters ($F_1^l$ and $F_2^l$ values) in the convolution kernels, so that we can have $\lambda_{l}^{t}$ close to $0$. The optimized ResNet50 architecture and the standard ResNet50 architecture are presented in Table \ref{tab:structure}. From a total of $23,534,467$ trainable parameters to $128,062$ trainable parameters, our method has demonstrated a significant reduction in memory and computational cost, as only $0.54\%$ of the parameters are kept after optimization. The performance remains unchanged after network structure optimization. More detailed information in training the network and result comparison are presented in the next section.
 
\section{Network Training and Evaluation}\label{sec:Evaluation and Visualization}
\subsection{Dataset and training process}
The public chest X-ray (CXR) dataset \citep{karargyris2021creation} shared through PhysioNet \citep{moody2000physionet} is used for the evaluation of our method. The chest X-ray dataset contains $1083$ CXR images with expert annotation. A CXR image is from a patient that is either healthy, has cardiomegaly (enlarged heart), or has pneumonia. The eye-gaze behaviors of the medical practitioner when examining the images are also recorded which can further assist with the deep learning and human-computer interaction algorithms \citep{zhu2022jointly, zhu2022gaze, zhu2019novel}. $70\%$ of the dataset is used for training; $10\%$ of the dataset is used for validation to avoid overfitting; $20\%$ of the dataset is used for evaluation. Similar to the method proposed by \citeauthor{zhu2022gaze}, the ResNet50 CNN model is used as a benchmark for CXR image classification (identify if a patient is healthy, has pneumonia, or has cardiomegaly). To optimize the ResNet50's structure, we start by training with the standard network. Based on the CKRM values from three independently trained networks, we increase or decrease the convolution kernel dimensions. We then repeat the training process until $\lambda_l^t$ has a desired value close to $0$ for all $l$ values in the network. An empirical study shows that hyper-parameters (e.g., learning rate) used for the standard ResNet50 usually yield optimal training outcomes for the optimized ResNet50 architecture.
\par
To test the generalizability of our method, we also integrate our method with the GG-CAM method proposed in \citep{zhu2022gaze}. To sum up, four models are compared, which are standard ResNet50, optimized ResNet50, standard ResNet50 with GG-CAM, and optimized ResNet50 with GG-CAM, abbreviated as ResNet50-std, ResNet50-opt, GG-CAM ResNet50-std, and GG-CAM ResNet50-opt, respectively.
 
\subsection{Result evaluation}
\begin{figure}[t]
     \centering
     \subfloat[][AUC metrics]{\includegraphics[width=0.32\textwidth]{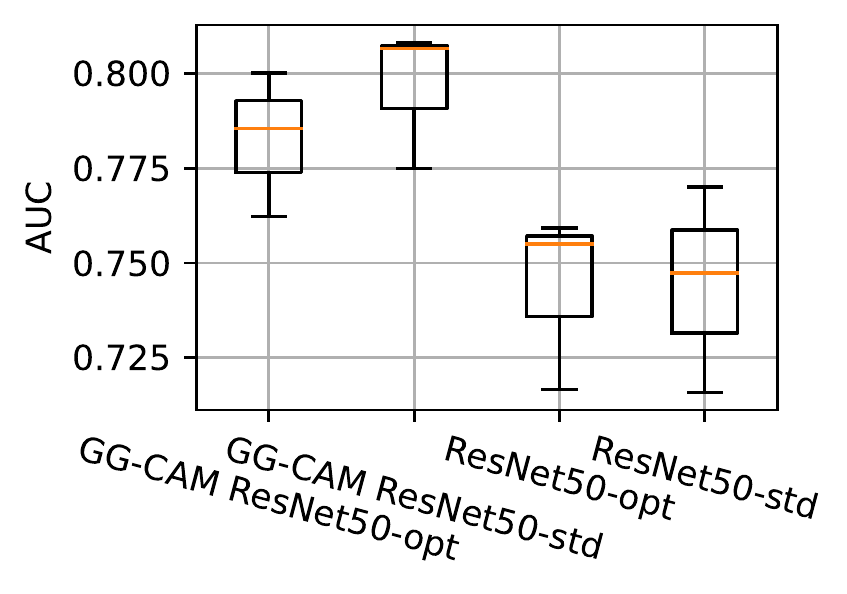}\label{fig:ACC}}
     \hfill
     \subfloat[][Accuracy metrics]{\includegraphics[width=0.32\textwidth]{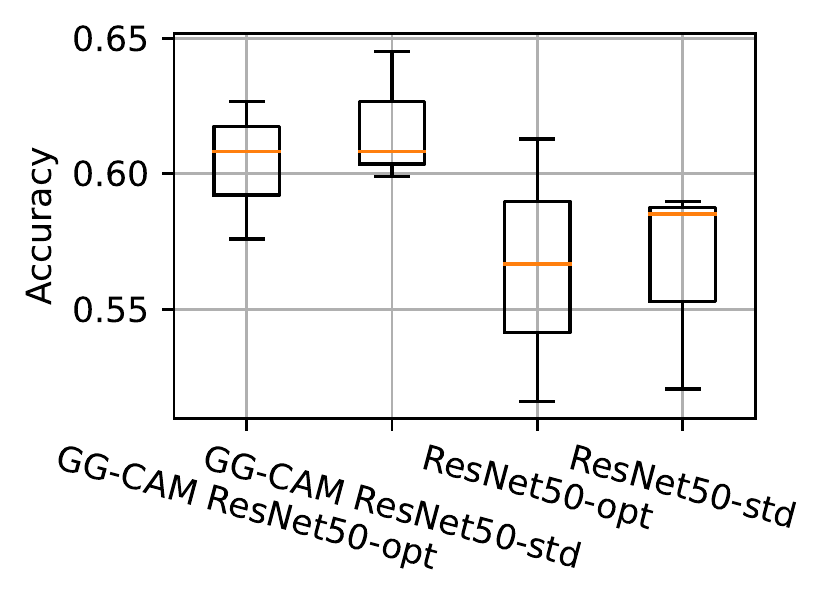}\label{fig:AUC}}
     \caption{Network performance comparisons}
     \label{fig:performance}
\end{figure}

To evaluate the performance of the networks, two metrics are used: the area under curve (AUC) for the receiver operating characteristics curve, and the accuracy for classification. Figure \ref{fig:performance} compares the results. We can see that ResNet50-std and ResNet50-opt yield comparable performances, validated with one-way ANOVA tests. For GG-CAM ResNet50-opt and GG-CAM ResNet50-std, as expected, GG-CAM boosts the classification performance. ANOVA test also confirms that the performances for GG-CAM ResNet50-opt and GG-CAM ResNet50-std are statistically equivalent. 
 
\section{Conclusion}\label{sec:Conclusion}
In this paper, we propose an effective network structure optimization method based on the assumption that we usually do not need as many kernels as defined in the standard networks for the extraction and processing of the information required by a designated task. Inspired by the SSIM method that is frequently used for comparing scenic image similarity, we adopt it for the evaluation of convolution kernel redundancy which are almost visually intractable other than the first convolution layer. The quantitative measurement of kernel redundancy lays the foundation for guided network structure optimization, and is exponentially more efficient than the randomized hyper-parameter search method. As performance of the optimized network is statistically equivalent with the standard structures, we have demonstrated the effectiveness of our method.
\par
Still, there are several limitations with our method that may require future improvement. One limitation is the lack of testing over a broader range of computer vision tasks. Future work can be done to evaluate how our method performs when we change the underlying task, or have another CNN architecture. Also, our method neglects the inter-layer interaction. Changing the kernel dimension of a shallower convolution layer will impact the information flow to deeper layers, which cannot be captured with our method. Further improvement on the optimized network structure may be possible if we take inter-layer interaction into consideration.

\bibliography{mybib}
 
 
\end{document}